\RequirePackage{fix-cm}
\documentclass[smallcondensed]{svjour3}     
\smartqed  
\usepackage{booktabs}
\usepackage{natbib}
\usepackage{amsfonts}
\usepackage{graphicx}
\usepackage{algorithm2e}
\usepackage{multirow}
\SetKwInput{kwSym}{Notation}
\SetKwInput{kwCalc}{Calculate number of instances to oversample}

\usepackage{caption}
\usepackage{subcaption}

\usepackage{wrapfig}
\usepackage{float}
 \journalname{Machine Learning}
\begin{document}

\title{Towards A Holistic View of Bias in Machine Learning: Bridging Algorithmic Fairness and Imbalanced Learning
}
\subtitle{}

\titlerunning{Towards A Holistic View of Bias in Machine Learning}        

\author{Damien A. Dablain \and Bartosz Krawczyk \and Nitesh V. Chawla}


\institute{Damien A. Dablain and Nitesh V. Chawla \at
 Department of Computer Science and Engineering and the Lucy Family Institute for Data and Society\\
 University of Notre Dame\\
 Notre Dame, IN 46556 USA
           \and
  Bartosz Krawczyk \at
Department of Computer Science\\
Virginia Commonwealth University\\
Richmond, VA 23284 USA
}


\maketitle

\vspace{-1cm}

\begin{abstract}
Machine learning (ML) is playing an increasingly important role in rendering decisions that affect a broad range of groups in society. ML models inform decisions in criminal justice, the extension of credit in banking, and the hiring practices of corporations. This posits the requirement of {\it{model fairness}}, which holds that automated decisions should be equitable with respect to protected features (e.g., gender, race, or age) that are often under-represented in the data. We postulate that this problem of under-representation has a corollary to the problem of imbalanced data learning. This class imbalance is often reflected in both classes and protected features. For example, one class (those receiving credit) may be over-represented with respect to another class (those not receiving credit) and a particular group (females) may be under-represented with respect to another group (males). A key element in achieving algorithmic fairness with respect to protected groups is the simultaneous reduction of class and protected group imbalance in the underlying training data, which facilitates increases in both model accuracy and fairness. We discuss the importance of bridging imbalanced learning and group fairness by showing how key concepts in these fields overlap and complement each other; and propose a novel oversampling algorithm, \emph{Fair Oversampling}, that addresses both skewed class distributions and protected features. Our method: (i) can be used as an efficient pre-processing algorithm for standard ML algorithms to jointly address imbalance and group equity; and (ii) can be combined with fairness-aware learning algorithms to improve their robustness to varying levels of class imbalance. Additionally, we take a step toward bridging the gap between fairness and imbalanced learning with a new metric, \emph{Fair Utility}, that combines balanced accuracy with fairness. Our source code and data are publicly available at https://github.com/dd1github/Fair-Over-Sampling.
\end{abstract}

\section{Introduction}
 Automated decision-making models are progressively being used in situations that affect humans in a broad range of areas, such as credit risk analysis \citep{khandani2010consumer}, criminal recidivism prediction \citep{berk2021fairness,chouldechova2018case}, hiring \citep{schumann2020we} and the provision of social services \citep{shroff2017predictive}. For example, a bank may decide to extend credit based on whether a machine learning (ML) model predicts that an individual may default on a loan. Conversely, a judge may determine that a defendant should not be released while awaiting trial if an artificial intelligence (AI) model suggests that the defendant has a high risk of recommitting a crime. The growing prevalence of ML algorithms in decisions that affect humans is due, in part, to their perceived accuracy and ability to detect hidden patterns in data. Yet, in some cases, these models have been demonstrated to incorporate biases, such as in hiring decisions \citep{reuters}, face recognition \citep{raji2020saving} and even translation \citep{caliskan2017semantics}, resulting in concerns about the fairness of machine learning algorithms \citep{barocas2016big}. Despite these concerns, it is likely that the use of automated decision-making will only increase in the future, as AI becomes more wide-spread in society, government and business. Therefore, there is a growing awareness that ML algorithms should be \emph{both} accurate and fair, which is underscored by the recently released Artificial Intelligence Act - a legal framework promulgated by the European Commission that mandates non-discrimination, among other requirements, for ML models that affect individuals \citep{eu}.



 
 Although it is desirable for ML models to be both fair and accurate, there is often a trade-off between these two goals \citep{kleinberg2018inherent,flores2016false,berk2021fairness}, such that increasing fairness comes at the cost of reduced accuracy. Accuracy and fairness are often at odds because they are influenced by imbalanced data. In many cases, the data used to train ML algorithms is imbalanced with respect to class and protected features, such that one class or group is under-represented with respect to another. Since most ML models learn parameters based on data, data imbalance can cause a particular class or sub-group to be over-weighted, such that preference is given to the over-represented class or group. Several previous works on algorithmic fairness do not explicitly take into account the relative sizes and imbalance of classes and protected groups \citep{dwork2012fairness,hardt2016equality,zafar2017fairness}, even though these characteristics can affect accuracy and equitable outcomes. Hashimoto et al. \citep{hashimoto2018fairness} has referred to the under-representation of a protected group in training data as \emph{representation disparity}, such that minority groups contribute less to a ML model objective because they are under-represented in the training data, and hence model accuracy may be lower for the minority class.
 
 \smallskip
 \noindent \textbf{Goal.} To draw out the corollary between algorithmic fairness and learning from imbalanced data and introduce a novel fairness-aware oversampling approach that addresses fairness within the construct of learning from imbalanced data. 
 
 \smallskip
 \noindent \textbf{Summary.} The algorithmic fairness domain focuses on combating bias in decision-making originating in protected features that could affect the objectiveness of the decision and lead to unfavorable results for under-represented and minority groups. At the same time, the class imbalance domain focuses on countering bias originating from skewed class distributions, as majority classes may be preferred over minority ones during classifier training. We postulate that when handling protected features to achieve fairness, one will always face some kind of imbalance in the training data. Therefore, algorithmic fairness and class imbalance should be seen as two sides of the same coin. Using techniques from the  imbalanced data domain can benefit algorithmic fairness, leading to the design of more robust and holistic methods that can counter both data and algorithmic biases. We take a step toward bridging the gap between algorithmic discrimination and imbalanced learning by discussing the key concepts and metrics that underpin both areas. Because it is often not possible for a ML algorithm to meet multiple fairness criteria (e.g., individual and group non-discrimination) at the same time \citep{chouldechova2017fair,corbett2017algorithmic}, we focus on a single element - group fairness. We show that common approach used in imbalanced learning - data oversampling - can be used to increase model fairness and accuracy. 
  
\smallskip
\noindent \textbf{Main contributions.} This paper offers the following insights and contributions towards fair and robust machine learning:
  
\begin{itemize}
\item{\textbf{Bridging the Gap Between Fairness and Imbalanced Learning}}: We take a step toward bridging the gap between the algorithmic fairness and imbalanced learning fields by showing that unbalanced training datasets affect a classifier's ability to produce equitable treatment of protected groups. Training sets that are proportionately balanced with respect to classes and protected groups result in fairer results when measured by three measures of group fairness - average odds, equalized odds and the difference in true negative rates.
\item {\textbf{Fair Oversampling}}: We propose a new data pre-processing technique, Fair Oversampling (FOS), that enhances fairness and classifier accuracy. The method is based on a popular imbalanced learning technique that is adaptable to group fairness. It is straight-forward and can be easily inserted into a fairness-aware ML model or standard classifier pipeline, improving robustness under varying levels of class imbalance.
\item{\textbf{Fair Utility metric}}: We propose a new metric that combines fairness with imbalanced learning - Fair Utility - that relies on measures commonly used in both fields. 
\end{itemize}

  \smallskip
  \noindent \textbf{Organization.} The paper is organized as follows. It first discusses algorithmic fairness and its key concepts. Then, it reviews the central elements of imbalanced learning. Next, it introduces our algorithm and proposed fairness metric. Finally, the paper discusses experimental results, and the commonalities between algorithmic fairness and imbalanced learning.

\section{Algorithmic fairness}
Discrimination can generally be defined as the prejudicial treatment of an individual based on membership in a legally protected group. 
Algorithmic fairness is concerned with ensuring that decisions made by machine learning models are equitable with respect to protected groups \citep{romei2014multidisciplinary,vzliobaite2017measuring}. Algorithmic fairness commonly invokes one of the following changes in order to produce equitable results: (1) modifications to the training data, (2) changes in the machine learning model, or (3) modifications to the decisions themselves \citep{calmon2017optimized}.

We concentrate on supervised learning in the context of binary classification. In binary classification, the goal of algorithmic fairness is to fairly select between two actions, $a_0$ and $a_1$ (e.g., approve or decline the extension of credit in banking). In our discussion, we adopt notation used by Speicher et al. \citep{speicher2018unified} to describe our algorithmic environment. Thus, a ML decision algorithm, \emph{A}, can be described as a function $A: \mathbb{R} \rightarrow \{0,1\}$ that outputs a binary decision. The machine learning algorithm, \emph{A}, parameterized by $\theta$, accepts as input training data, \emph{D}, minimizes a loss function $l(\theta)$, and predicts a label (i.e., 0 or 1). 

More formally, \emph{A} accepts as input training data, $D=\{(x_i,y_i)\}^n_{i=1}$, with \emph{n} examples, features $x_i \in X$, where $y_i \in Y$ represents the prediction or label ($Y = \{0,1\}$) for each individual \emph{i}. The features, \emph{X}, can either be discrete or continuous. We partition the set of features (or attributes) into two groups, sensitive or protected features, such as gender, race or age, and unprotected features, such that $x = (x_p, x_u)$. We also assume that protected features can be further partitioned into two classes, privileged and unprivileged (i.e., $x_p = (x_{pr}, x_{up})$).
For purposes of this paper, we assume that the label contained in the dataset is the correct, unbiased label.

Narayanan described at least 21 mathematical definitions of fairness that have been proposed by the fairness research community \citep{narayanan2018translation}. Two broad classes of algorithmic equity have gained prominence: group and individual fairness. Group fairness requires that the ML algorithm, \emph{A}, produces parity for a given metric, \emph{M}, for protected features, such that $M_{x_p}(A) = M_{x_p'} \forall\ x_p, x_p' \in X$. Individual fairness requires that similar individuals are treated similarly. It implies the presence of a similarity metric that is capable of determining if a pair of individuals are similar. The leading definition of individual fairness is metric fairness, as proposed by Dwork et al. \citep{dwork2012fairness}. It requires: $M_y(A(x_1),A(x_2)) \leq LM_x(x_1,x_2)\ \forall\ x_1, x_2 \in X,$ where $M_x$ and $M_y$ are data and label metrics, $A$ is a map or algorithm, and $L \geq 0$ is a Lipschitz constant. Finding such a similarity metric can be challenging. To overcome this obstacle, some works have assumed the existence of a publicly available similarity metric issued by a regulatory body that follows Rawlsian principles \citep{rawls2001justice,dwork2012fairness}.

Because of the challenges in finding suitable individual fairness similarity metrics, we focus on group fairness in this paper. Corbett-Davies et al. describe three central concepts embodied by group fairness: anti-classification, classification parity and calibration \citep{corbett2018measure}. Anti-classification requires that AI algorithms do not consider protected features when making decisions \citep{bonchi2017exposing,caliskan2017semantics,grgic2016case}. Thus, anti-classification provides that: $A(x) = A(x')\ \forall\ x$, such that $x_u = x_u'$. Classification parity (sometimes referred to as statistical parity) requires that certain measures are equal across sensitive features. Statistical parity can be expressed in a variety of ways. Under one formulation, the proportion of members in a protected group receiving a positive classification must be identical to the proportion in the population as a whole \citep{zemel2013learning}. Other measures focus on the difference in positive or negative rates (instead of proportions) between sensitive groups (e.g., equal true positive rates for both male and female applicants). Classification parity has been widely used as a fairness metric in machine learning \citep{agarwal2018reductions,calders2010three,edwards2015censoring}. As discussed below, we use classification parity in our metrics. Demographic parity, or the proportion of positive decisions, means that $Pr(A(X)=1 | x_p) = Pr(A(X)= 1)$ \citep{feldman2015certifying}. Whereas, parity of false positives requires that $Pr(A(X) = 1 | Y = 0, x_p) = Pr(A(X) = 1 | Y = 0)$. We also incorporate demographic parity into our metrics, although we focus on differences in true positive, false positive, and true negative rates, instead of their relative proportions.

Group fairness concepts are firmly entrenched in U.S. society and law. For example, the Fourteenth Amendment to the U.S. Constitution contains an equal protection law that prevents government workers from acting with "discriminatory purpose" against protected groups. Laws issued by the US Supreme Court also prevent discriminatory actions against protected groups, as formalized by \emph{Griggs v. Duke Power Co.}. The ubiquity of group fairness concepts in society is another reason why we focus on them in our analysis.

\smallskip
\noindent\textbf{Fairness-aware algorithms.} In order to achieve group fairness in machine learning, a variety of techniques have been employed, which can be broadly separated into pre-processing, in-processing and post-processing methods. Pre-processing techniques involve manipulating the training data before it is consumed by a classification algorithm, in-processing incorporates fairness into a ML algorithm loss function, and post-processing aims to adjust the decisions of a classifier to be fair. We briefly survey below the key pre-processing and in-processing techniques that are relevant to our approach.

\smallskip
\noindent\textbf{Pre-processing techniques.} Kamiran and Calders propose a pre-processing method, Reweighing, that creates weights for the training instances to ensure fairness \citep{kamiran2013quantifying}. They effectively divide the training set into four groups: (1) privileged group, majority class; (2) unprivileged group, majority class; (3) privileged group, minority class; and (4) unprivileged group, minority class. They then develop separate weights for each of the four groups and apply the weights to each instance. Feldman et al. propose a pre-processing method, Disparate Impact Remover, that modifies features to enhance group fairness while preserving rank-ordering within protected groups \citep{feldman2015certifying}.

\smallskip
\noindent\textbf{In-processing techniques.} Agarwal et al. propose an in-processing method, Exponentiated Gradient Reduction (EGR), that reduces fair classification to a sequence of cost-sensitive classification problems, whose solution yields a randomized classifier with the lowest error subject to the desired constraints 
\citep{agarwal2018reductions}. Zhang et al. propose Adversarial Debiasing to mitigate unwanted biases \citep{zhang2018mitigating}. Their method uses adversarial training to prevent an adversary from determining a protected attribute from a ML model's predictions. This approach results in a fair classifier because the predictions are not allowed to include group discrimination information. Yurochkin, Bower and Sun propose a method, Sensitive Subspace Robustness (SSR), based on adversarial training \citep{yurochkin2019training}. SSR induces individual fairness based on sensitive perturbations of inputs. It casts fairness in the form of robustness to sensitive perturbations of the training data. SSR uses a fair Wasserstein distance metric to require the output of a ML model to be similar to the training label. 

\section{Imbalanced learning}

Imbalanced learning is concerned with disproportions among classes. In binary classification, the number of instances of one class (the majority) outnumber the other (minority). The skewed distribution of examples in favor of the majority class can cause classifiers to be biased toward the majority because the algorithm's parameters are more heavily weighted toward more frequently occurring examples. Classifiers can achieve high accuracy by merely selecting the majority class. However, the minority class is often the more important one from the data inference perspective because it may carry more relevant information. For example, in anomaly detection, there may be few examples of malicious software, which is the more interesting class. In addition to detecting malicious software in computer security \citep{cieslak2006combating}, imbalanced learning has been applied to fraud detection \citep{wei2013effective}, medical treatment (i.e., identifying rare diseases) \citep{rao2006data}, image recognition \citep{kubat1998machine}, atypical behavior in social networks and infrastructure monitoring systems \citep{krawczyk2016learning}. Training sets where one class is under-represented compared to the majority class frequently results in poor accuracy performance by the classifier \citep{seiffert2007mining}.

\smallskip
\noindent \textbf{Taxonomy.} There are three broad approaches within imbalanced learning: data-level methods that modify the training data to balance class distributions, algorithm level methods that ameliorate bias in classifiers towards the majority class, and ensemble methods that are a combination of the first two with classifier committees.

\smallskip
\noindent \textbf{Data-level approaches.} This group of methods focus on modifying the training set by balancing the number of minority and majority class examples. Oversampling generates new minority class examples, while under-sampling removes instances from the majority class. Under-sampling can result in removal of important data from the training set and therefore is often not preferred. Simple random oversampling (ROS) merely duplicates instances of the minority class to impose parity. SMOTE, or the Synthetic Minority Oversampling Technique, \citep{chawla2002smote}, is a popular oversampling method used in the imbalanced learning community. It randomly selects a nearest neighbor of a minority instance and linearly generates synthetic examples based on the original instance and a nearest neighbor. SMOTE has been adapted to enhance the importance of class borderline instances \citep{han2005borderline}, define safe regions that do not sample from noisy or overlapping instances \citep{bunkhumpornpat2009safe} and has been applied in the deep learning \citep{dablain2021deepsmote} and big data \citep{Sleeman:2021} contexts. Alternative approaches to SMOTE have been proposed recently that do not rely on $k$-nearest neighbors, instead using alternative measures such as class potential \citep{Krawczyk:2020}, Mahalanobis distance \citep{Sharma:2018}, or manifold approximation \citep{Bej:2021}. 

\smallskip
\noindent \textbf{Algorithm-level approaches.} This group of methods modify the training procedure of a classifier to make it skew-insensitive, or incorporate alternative cost functions. Cost sensitive learning, which is a form of importance sampling \citep{kahn1953methods}, magnifies the importance of minority examples by increasing the penalty associated with the instances. Recent examples of cost-sensitive methods that have been used in imbalanced learning include the focal loss \citep{lin2017focal}, the class-balanced margin loss \citep{cui2019class}, the distribution aware margin loss \citep{cao2019learning} and the asymmetric loss \citep{ridnik2021asymmetric}. 

\smallskip
\noindent \textbf{Ensemble approaches.} Combining multiple classifiers is considered as one of the most effective approaches in modern machine learning \citep{wozniak2014survey}. Ensembles find their natural application in learning from imbalanced data, as they leverage the predictive power of multiple learners. By combining base classifiers with data or algorithm-level solutions, they achieve locally specialized robustness and maintain diversity among ensemble members. Most popular solutions combine resampling with Bagging \citep{Lango:2018} or Boosting \citep{Zhang:2019wot}, use mutually complimentary cost-sensitive learners \citep{Tao:2019}, or rely on dynamic selection mechanisms to tackle locally difficult decision regions \citep{Zyblewski:2021}. 

\section{Why we need to bridge algorithmic fairness and imbalanced learning?}

\noindent \textbf{Different views on bias.} The two previous sections provided general background and reviewed recent advancements in algorithmic fairness and imbalanced learning. This allows us to see the strong parallel between them, as they both deal with the problem of countering bias in data and algorithms, however from different perspectives:

\begin{itemize}
\item \textbf{Bias according to algorithmic fairness.} Here, bias is seen as a lack of fairness and transparency, originating from social background and the nature of the data itself. Fairness focuses on bias based on using sensitive or protected information (e.g., race or gender) to make a decision, thus putting traditionally under-represented groups at a significant disadvantage. Fairness-aware algorithms also focus on using safe information for training classifiers and debiasing them with respect to protected features.
\item \textbf{Bias according to imbalanced learning.} Class imbalance focuses on bias originating in disproportion among classes, as most machine learning algorithms will become biased towards classes with a higher number of training instances. This puts smaller, yet often more important, classes at a disadvantage. Imbalance-aware algorithms focus on either balancing class distributions or removing the bias towards majority classes from the training process.
\end{itemize} 

\noindent \textbf{Two sides of the same coin.} We can see that algorithmic fairness views the source of bias as the information being used to make a decision, while class imbalanced learning views bias as arising from a disproportion among instances used to train a classifier. In this paper, we argue that these are two sides of the same coin. While bias in decision making may stem from using sensitive features, they are usually followed by imbalanced distributions, as traditionally under-represented groups are akin to minority classes. Therefore, considering only one type of bias leads to an oversimplified view of the problem. To achieve a truly fair and robust machine learning algorithms, we need to develop a  holistic view where both types of bias are tackled during classifier training.

\section{Fair Oversampling}

Our algorithm, Fair Oversampling (FOS), is designed to improve fairness and increase classifier accuracy. The basic intuition behind FOS is that bias in machine learning models is caused, in part, by under-represented classes and features in training sets.  When training a machine learning model to accurately and fairly discern classes and protected features, it is often necessary to equalize the \emph{number} of training examples between classes and protected groups to ensure that algorithmic models based on parametric learning are able to balance gradients and model weights between specific classes and features.  If a classifier observes very few instances of a minority class or certain minority features, it's parameters may be biased toward recognizing the dominant class or group.  FOS recognizes that numerical class and feature imbalance may exist in the data used to train machine learning models and aims to redress this numerical data imbalance - thus enhancing both model predictive accuracy and fairness.

As a pre-processing algorithm, it modifies a training dataset $D$ so that it can be input to the machine learning model. The modified data can either be used with a standard classifier, such as Support Vector Machines (SVM), or with an algorithm that is specifically designed to improve fairness, such as SSR or Adversarial Debiasing. FOS acts on two types of independent variables (X) in the training data, protected features $x_p$, which are binary, and unprotected features $x_{u}$, which are expressed as real numbers. The pseudocode for FOS is displayed in Algorithm~\ref{alg:algo}.

\RestyleAlgo{ruled}
\begin{algorithm}[]
\scriptsize
\caption{Fair Oversampling Pseudocode}\label{alg:algo}
\kwSym{\\
$D$ - training dataset;\\
$L$ - training labels;\\
$D_{prmaj}$ - privileged group, majority class; \\ 
$D_{upmaj}$ - unprivileged group, majority class; \\ 
$D_{prmin}$ - privileged group, minority class; \\ 
$D_{upmin}$ - unprivileged group, minority class; \\ 

$N_{prmaj}$ - number of $D_{prmaj}$; \\ 
$N_{upmaj}$ - number of $D_{upmaj}$; \\
$N_{prmin}$ - number of $D_{prmin}$; \\
$N_{upmin}$ - number of $D_{upmin}$; \\

$S_{pr}$ = number of $D_{prmin}$ to oversample; \\ 
$S_{up}$ = number of $D_{upmin}$ to oversample;\\}
\kwCalc{}
$S_{pr}$ = $N_{prmaj}$ - $N_{prmin}$; \\ 
$S_{up}$ = $N_{upmaj}$ - $N_{upmin}$;\\
\eIf{$S_{up}$ is less than $S_{pr}$}{
 $N_{samp1} = S_{up}$\; 
 $N_{samp2} = S_{pr}$\;
 $D_1 = D_{upmin}$\;
 $D_2 = D_{prmin}$\;
 }{{
 $N_{samp1} = S_{pr}$\;
 $N_{samp2} = S_{up}$\;
 $D_1 = D_{prmin}$\;
 $D_2 = D_{upmin}$\;
 }
 }
$Base_1 \gets randomly\ select\ N_{samp1}\ from\ D_1$\;
\For{$B\ in\ Base_1$}{
 $KNN \gets select\ K\ nearest\ neighbors\ of\ B\ from\ D_1$\;
 $N \gets randomly\ select\ 1\ neighbor\ from\ KNN$\;
 $R \gets randomly\ select\ a\ number\ between\ 0\ and\ 1$\;
 $S = B\ +\ R\ X\ (B\ -\ N)$\;
 $Samples \gets append\ S$\;
 $Labels \gets append\ label\ of\ D_1$\;
 }
$X \gets D\ +\ Samples$\;
$Y \gets L\ +\ Labels$\;

$Base_2 \gets randomly\ select\ N_{samp2}\ from\ D_2$\;
\For{$B\ in\ Base_2$}{
 $KNN \gets select\ K\ nearest\ neighbors\ of\ B\ from\ D_2$\;
 $N \gets randomly\ select\ 1\ neighbor\ from\ KNN$\;
 $R \gets randomly\ select\ a\ number\ between\ 0\ and\ 1$\;
 $S = B\ +\ R\ X\ (B\ -\ N)$\;
 $Samples \gets append\ S$\;
 $Labels \gets append\ label\ of\ D_1$\;
 }
$X \gets X\ +\ Samples$\;
$Y \gets Y\ +\ Labels$\
\end{algorithm}

FOS first determines the minority and majority classes ($Y = \{0,1\}$ or $Y = \{min,maj\}$) . It then subdivides the protected features $x_p$ into two categories - privileged and unprivileged ($x_p = (x_{pr}, x_{up})$). For example, in a dataset $D$ related to the extension of credit, the majority class ($D_{maj}$) could be people that receive credit, and the minority class ($D_{min}$) could be those that do not receive credit. The protected feature $x_p$ could be gender, where males are considered privileged $x_{pr}$ and females are unprivileged $x_{up}$. This categorization results in four sub-groups: privileged majority ($D_{prmaj}$), unprivileged majority ($D_{upmaj}$), privileged minority ($D_{prmin}$) and unprivileged minority ($D_{upmin}$).

The objective of FOS is to restore balance between the classes and protected features through random oversampling and nearest neighbor metrics, using the mechanics of the SMOTE algorithm, but with modifications. FOS numerically balances the classes and protected features, such that the number of examples ($N$) in the majority class ($N_{maj}$) equals the number of examples in the minority class ($N_{min}$), or $N_{maj}$=$N_{min}$.  It determines the protected group $x_p$ in the dataset $D$ that requires the least number of samples to obtain equivalency (denoted as $D_1$), and selects the K nearest neighbors of a random sample of $D_1$ (e.g., the unprivileged, minority group $D_{upmin}$). In our implementation, five nearest neighbors were selected; however, K selection can depend on the specific dataset. The number of random samples selected from this group equals the number of samples required to make it equal in number to the same group in the majority class (e.g., $N_{upmin}$=$N_{upmaj}$). For $D_1$, the samples are drawn from a single protected sub-group (e.g., the samples are exclusively drawn from $D_{upmin}$). 

Next, the same oversampling procedure is repeated for the protected group $x_p$ with the larger number of samples that are required to obtain numerical equivalency (e.g., $D_{prmin}$), which is denoted as $D_2$, except that instead of drawing the nearest neighbors exclusively from the $D_2$ pool, they are drawn from the entire minority class. For example, the samples are drawn from $D_{min}$ instead of $D_{prmin}$.   Empirically, we found that this approach reduced bias because it blurs the difference between privileged minority $D_{prmin}$ and unprivileged minority $D_{upmin}$ group members, since it draws a nearest neighbor from the entire minority class $D_{min}$, which consists of both privileged and unprivileged members. This approach provides an opportunity to effectively increase the representation of privileged, minority $D_{prmin}$ members if the features of a privileged minority $D_{prmin}$ member are nearest in proximity to an unprivileged, minority member $D_{upmin}$.  Our sampling procedure is different than the SMOTE approach, which selects nearest neighbors solely from the combined minority class, without regard to protected groups.

FOS balances the \emph{number} of class and protected group instances within a training dataset.  It does not balance protected attribute \emph{ratios}.  We postulate that FOS produces fair results, even without balanced protected group ratios, because it acts on the weights of parametric machine learning models. By allowing the machine learning models to observe an equal \emph{number} of class and protected group instances, it facilitates balanced parametric learning (e.g., equal weighted average number of instances), even though the natural ratio of feature instances may vary. 

\section{Experiments}

The following experiments were designed to answer the following research questions:

\begin{itemize}
    \item[RQ1:] Does FOS improve both algorithmic fairness and robustness to class imbalance for popular standard classifiers?
    \item[RQ2:] Can FOS further improve the performance of fairness-aware classifiers?
    \item[RQ3:] What is the FOS trade-off between fairness and skew-insensitive metrics when handling varying imbalance ratio levels?
    \item[RQ4:] How does FOS influences feature importance used by underlying the classifiers and does it lead to more fairness-aware feature selection?
\end{itemize}

\subsection{Datasets} 

Three popular datasets were selected for testing that were used by the fairness research community \citep{Quy:2021}: German Credit \citep{uciger}, Adult Census Income \citep{kohavi1996scaling}, and Compas Two-Year Recidivism \citep{propublic}. The key statistics of each dataset are summarized in Table~\ref{tab: ddesc}. All three datasets involve binary classification. The German Credit dataset contains data that allows a classifier to predict whether an individual should have a positive or negative credit rating. The Adult Census Income dataset predicts whether an individual earns more or less than \$50K. The Compas dataset can be used to predict whether a defendent will commit a crime within a two year period. 

As we can see from Table~\ref{tab: dimbal}, all of the datasets exhibit both class and protected attribute (gender) imbalance. Compas shows the least amount of class imbalance, with a ratio of 1.22:1, while the German Credit and Adult Census datasets have class imbalance ratios ranging from approximately 2:1 to 3:1. All three datasets show greater protected attribute imbalance than class imbalance, with the ratios ranging from approximately 2:1 to 4:1. In the minority class, the maximum protected attribute imbalance ratios are even higher, ranging from 1.75:1 in German Credit to 5.61:1 in Adult Census.

\begin{table}[!h]
\centering
\footnotesize
\caption{Description of the Datasets}
 \label{tab: ddesc}
\begin{tabular}{llcp{1.2cm}l}
\toprule
\textbf{Dataset}& \textbf{Instances} & \textbf{Features} &\textbf{Protected Feature}& \textbf{Classes} \\
\midrule
German Credit & 1,000 & 20 & Gender & Good credit; bad credit \\
Adult Census Income & 48,842 & 14 & Gender & Income $>$ or $<$ 50K\\
Compas & 7,214 & 28 & Gender & Recidivism; No recidivism  \\
\bottomrule
\end{tabular}
\end{table}

\begin{table}[!h]
\footnotesize
\centering
\caption{Class and protected feature imbalance ratios for each dataset}
 \label{tab: dimbal}
\begin{tabular}{llll}
\toprule
\textbf{Class / Feature}& \textbf{German Credit} & \textbf{Adult Census} &\textbf{Compas} \\
\midrule
\multicolumn{4}{l}{\textit{Classes}} \\
\midrule
Majority & 700 & 37,155 & 3,963\\
Minority & 300 & 11,687 & 3,251\\
Ratio & 2.33 & 3.18 & 1.22\\
\midrule
\multicolumn{4}{l}{\textit{Protected Features}} \\
\midrule
Privileged & 690 & 32,650 & 5,819\\
Unprivileged & 310 & 16,192 & 1,395\\
Ratio & 2.23 & 2.02 & 4.17\\
\midrule
\multicolumn{4}{l}{\textit{Combined Class and Protected Features}}\\
\midrule
Privileged, majority & 499 & 22,732 & 3,066\\
Unprivileged, majority & 201 & 14,423 & 897\\
Ratio & 2.48 & 1.58 & 3.42\\
\midrule
Privileged, minority & 191 & 9,918 & 2,753\\
Unprivileged, minority & 109 & 1,769 & 498\\
Ratio & 1.75 & 5.61 & 5.53\\
\bottomrule
\end{tabular}
\end{table}

\subsection{Experimental design} 
\noindent \textbf{Experiment 1: Oversampling for standard classifiers.} First, modified training data produced by FOS was used as input to two standard machine learning classifiers: SVM and Logistic Regression (LG). The performance of the models was assessed based on metrics, which are discussed below. The performance of our algorithm was compared against four benchmarks for each standard classifier: (1) a baseline (no modifications to the training dataset); (2) a popular imbalanced learning oversampling method - SMOTE \citep{chawla2002smote}; and two pre-processing algorithms that are specifically designed to improve fairness - (3) Reweighing \citep{calmon2017optimized} and (4) Disparate Impact Remover \citep{feldman2015certifying}. The purpose of this experiment was to determine how FOS compared to other data pre-processing algorithms that are used in both the imbalanced learning and fairness research communities. 

\smallskip
\noindent \textbf{Experiment 2: Oversampling for fairness-aware classifiers.} Second, we assessed whether FOS could be used in conjunction with in-processing fairness-aware algorithms. For this experiment, the SSR \citep{yurochkin2019training}, Adversarial Debiasing \citep{zhang2018mitigating} and EGR models \citep{agarwal2018reductions}were chosen. To determine whether FOS could improve accuracy and fairness for these algorithms, the models were alternatively run with data that was not pre-processed and data that was modified by FOS.

\smallskip
\noindent \textbf{Experiment 3: Robustness to increasing imbalance ratios.} Third, we assessed how the performance of a standard ML classifier was affected by increasing levels of class and protected group imbalance. For this test, SVM was used as the ML algorithm with varying degrees of imbalance. Instances were randomly removed from classes and protected groups to achieve the intended imbalance levels. The selected imbalance levels were: $I \in \{{1,1.2,1.4,1.6,1.8,2\}}$ for German Credit; $I \in \{{1,2,2.5,3,3.5,4\}}$ for Compas; and $I \in \{{1,2,4,6,8,10\}}$ for Adult Census, where I represents the denominator of the fraction that reduced the number of original protected group members. The reason for different levels of imbalance by dataset is that classifiers trained with the Reweighing and Disparate Impact Remover algorithms produced unstable results for datasets with a relatively small number of examples (i.e., German Credit and Compas), such that the classifiers predicted all labels to reside in a single class, thus causing True Negatives to be zero, yielding "Nan" metrics. In contrast, both SMOTE and FOS were able to work at the $I = 10$ level for all datasets. Therefore, the imbalance ratio scaling was adjusted so that all pre-processing algorithms could be assessed for all datasets. 

\smallskip
\noindent \textbf{Experiment 4: Impact of oversampling on feature importance.} Fourth, we considered whether FOS caused a standard classifier to change the selection of the features that it used to formulate its decision boundary. For this purpose, the SVM ML algorithm was selected and feature importances of a classifier were compared using baseline training data and data modified by FOS to determine whether modifying the training data changed model feature selection.

\smallskip
\noindent \textbf{Setup.} All experiments were performed using five fold cross-validation. The reported results are averaged over the respective held out validation sets.  See Tables ~\ref{tab:svm}, ~\ref{tab:lg}, ~\ref{tab:fclass}.

\subsection{Metrics}
For purposes of the experiments, group fairness metrics were selected that could be expressed as elements of a binary classification confusion matrix consisting of True Positive Rate (TPR), False Positive Rate (FPR), True Negative Rate (TNR), and False Negative Rate (FNR). Metrics were chosen that are widely used in the fairness and imbalanced learning communities: Balanced Accuracy, Average Odds Difference (AOD), Absolute Average Odds Difference (AAO), Equal Opportunity Difference (EOD), and True Negative Rate Difference (TNRD) \citep{hardt2016equality,biswas2020machine,chakraborty2021bias}. AOD is the average difference in the False Positive Rate plus the True Positive Rate for privileged and unprivileged groups \citep{bellamy2018ai}. It can be expressed as: $\frac{1}{2} ((TPR_p - TPR_{up}) + (FPR_p - FPR_{up}))$, where $TPR_p$ is the TPR of privileged instances, $TPR_{up}$ is the TPR of the unprivileged instances, $FPR_p$ is the FPR of privileged instances, and $FPR_{up}$ is the FPR of the unprivileged instances. AAO is the same as AOD, except that TPR and FPR are absolute value calculations. EOD is the difference between the True Positive Rate of privileged and unprivileged groups \citep{bellamy2018ai}, and can be expressed as: $(TPR_p - TPR_{up}) + (FPR_p - FPR_{up})$. TNRD is $(TNR_p - TNR_{up})$, where $TNR_p$ is the TNR of privileged instances, and $TNR_{up}$ is the TNR of the unprivileged instances.

\subsubsection{Proposed metric.} In addition to the metrics discussed above, we propose a new metric, called {\bf{Fair Utility}}. In developing this metric, we are inspired by Corbett-Davies et al. \citep{corbett2017algorithmic}. They characterize algorithmic fairness in terms of constrained optimization in the context of the COMPAS algorithm for determining whether defendants in Broward County, FL, who were awaiting trial, were too dangerous to be released. In their formulation, the objective of algorithmic fairness is to \emph{both} maximize public safety and reduce racial disparities. We also view algorithmic fairness as a multi-objective optimization problem, where the goal is to maximize the accuracy of a classifier and reduce group inequality. We approach the optimization problem with a data pre-processing technique designed to balance class accuracy prediction with protected group equity. We are also inspired by Halevy, Norvig and Pereira, who postulated that, in machine learning, a large quantity of data is more important than a strong algorithm \citep{halevy2009unreasonable}. Our metric named \emph{Fair Utility} can be expressed as balanced accuracy multiplied by the average of TPRD plus TNRD. More explicitly, it is: $BA\ X\ \frac{1}{2}\ X\ ((1 - |TPRD|) + (1 - |FPRD|))$, where BA is balanced accuracy, TPRD is $(TPR_p - TPR_{up})$, and FPRD is $(FPR_p - FPR_{up})$. Utility involves maximizing the benefit of taking an action, compared with its costs. Here, we treat accuracy as equivalent with utility, which assumes that the class label assigned by the dataset is correct and does not contain inherent bias. The objective of {\it{Fair Utility}} is to combine accuracy and fairness into a single metric by incorporating balanced accuracy (which reflects the impact of class imbalance) with two fairness metrics (true positive and true negative rates) that track whether a classifier consistently accepts or rejects protected group members.

\subsection{Experiment 1: Oversampling for standard classifiers} 
FOS displays strong performance with respect to the standard classifiers, with clear improvements in fairness, as measured by AOD, AAO, TNRD, and EOD; although it shows better results with SVM as compared to Logistic Regression. For the SVM classifier, it consistently outperforms the other algorithms in terms of average odds, absolute average odds, TNRD, and Fair Utility. See Table~\ref{tab:svm}. It came in a close second to SMOTE in terms of Balanced Accuracy and clearly outperformed SMOTE in terms of the fairness metrics. Although SMOTE displays strong balanced accuracy, it often does not produce fair results with respect to protected groups.  This is likely because it balances the \emph{class} distribution, which improves the class false positive rate; while it is not designed to improve the false positive rate with respect to specific instance features. In terms of equal odds, FOS demonstrates significant reductions in unfairness compared to the baseline, with a first place finish for Compas and second place finishes for German Credit and Adult Census.

\begin{table*}
\footnotesize
 \caption{SVM Classifier Results}
 \label{tab:svm}
 \scalebox{0.8}{
 \begin{tabular}{lcccccc}
\toprule
 \textbf{Method}&\textbf{Bal Acc} &\textbf{Avg Odds} & \textbf{Absol Avg Odds}&\textbf{ TNRD}&\textbf{Equal Odds}&\textbf{Fair Utility}\\
\midrule
 \textbf{German Dataset}&&&&&&\\
 \texttt{Base} & 0.6486& 0.1054& 0.1054& 0.1358& 0.0750& 0.5795 \\
 \texttt{SMOTE}& \textbf{0.7055} & 0.0500& 0.0531& 0.0812& \textbf{0.0250}& 0.6677\\
 \texttt{Reweigh}& 0.6883& 0.0411& 0.0492& 0.0506& 0.0478& 0.6546\\
 \texttt{Disparate}& 0.6152& 0.0745& 0.0788& 0.1071& 0.0504& 0.5668\\
\midrule
 \texttt{FOS}& 0.7003& \textbf{0.0174}& \textbf{0.0313}& \textbf{0.0166}& 0.0460& \textbf{0.6781}\\
\midrule
 \textbf{Adult Dataset}&&&&&&\\
 \texttt{Base} & 0.7052& 0.1886& 0.1886& 0.0666& 0.3106& 0.5722 \\
 \texttt{SMOTE}& \textbf{0.8044} & 0.3094& 0.3094& 0.2934& 0.3255& 0.5527\\
 \texttt{Reweigh}& 0.7667& 0.0626& 0.0639& 0.1174& \textbf{0.0103}& 0.7177\\
 \texttt{Disparate}& 0.7259& 0.4378& 0.4378& 0.2687& 0.6070& 0.4035\\
\midrule
 \texttt{FOS}& 0.7935& \textbf{0.0208}& \textbf{0.0422}& \textbf{0.0247}& 0.0598& \textbf{0.7600}\\
\midrule
 \textbf{Compas Dataset}&&&&&&\\
 \texttt{Base} & 0.6641& 0.2022& 0.2022& 0.1541& 0.2503& 0.5296 \\
 \texttt{SMOTE}& \textbf{0.6700} & 0.2176& 0.2176& 0.1835& 0.2517& 0.5242\\
 \texttt{Reweigh}& 0.6447& 0.0319& 0.0353& 0.0462& 0.0244& 0.6220\\
 \texttt{Disparate}& 0.6613& 0.0486& 0.0500& 0.0593& 0.0407& 0.6280\\
\midrule
 \texttt{FOS}& 0.6680& \textbf{0.0125}& \textbf{0.0246}& \textbf{0.0267}& \textbf{0.0224}& \textbf{0.6512}\\
\bottomrule
 \end{tabular}
 }
\end{table*}

For Logistic Regression, FOS consistently produced the top Fair Utility results, with first place finishes in terms of average odds and absolute average odds, and first place misses by less than .0057 points. It also showed significant reductions in equal odds and TNRD, when compared to baselines, with first or second place results. See Table~\ref{tab:lg}.

\begin{table}
\footnotesize
 \caption{Logistic Regression Classifier Results}
 \label{tab:lg}
  \scalebox{0.8}{
 \begin{tabular}{lcccccc}
 \toprule
 \textbf{Method}&\textbf{Bal Acc} &\textbf{Avg Odds} & \textbf{Absol Avg Odds}&\textbf{ TNRD}&\textbf{Equal Odds}&\textbf{Fair Utility}\\
\midrule
 \textbf{German Dataset}&&&&&&\\
 \texttt{Base} & 0.6469& 0.0886& 0.1084& 0.0730& 0.1438& 0.5756 \\
 \texttt{SMOTE}& \textbf{0.7158} & 0.0659& 0.0867& 0.0688& 0.1046& 0.6536\\
 \texttt{Reweigh}& 0.6099& 0.0679& 0.0679& \textbf{0.0387}& 0.0970& 0.5683\\
 \texttt{Disparate}& 0.6385& 0.0648& 0.0793& 0.0417& 0.1169& 0.5875\\
\midrule
 \texttt{FOS}& 0.7144& \textbf{0.0452}& \textbf{0.0563}& 0.0414& \textbf{0.0712}& \textbf{0.6746}\\
\midrule
 \textbf{Adult Dataset}&&&&&&\\
 \texttt{Base} & 0.6765& 0.1960& 0.1960& 0.0726& 0.3193& 0.5439 \\
 \texttt{SMOTE}& \textbf{0.7595} & 0.3238& 0.3238& 0.3776& 0.2700& 0.5136\\
 \texttt{Reweigh}& 0.6941& \textbf{0.0080}& \textbf{0.0113}& \textbf{0.0056}& 0.0170& 0.6862\\
 \texttt{Disparate}& 0.6532& 0.0359& 0.0359& 0.0156& 0.0561& 0.6298\\
\midrule
 \texttt{FOS}& 0.7352& 0.0137& 0.0158& 0.0162& \textbf{0.0153}& \textbf{0.7236}\\
\midrule
 \textbf{Compas Dataset}&&&&&&\\
 \texttt{Base} & 0.6649& 0.2129& 0.2129& 0.2704& 0.1553& 0.5230 \\
 \texttt{SMOTE}& \textbf{0.6733} & 0.2271& 0.2271& 0.2578& 0.1963& 0.5200\\
 \texttt{Reweigh}& 0.6669& 0.0170& 0.0285& 0.0327& \textbf{0.0243}& 0.6476\\
 \texttt{Disparate}& 0.6675& 0.0844& 0.0844& 0.1027& 0.0661& 0.6111\\
\midrule
 \texttt{FOS}& 0.6674& \textbf{0.0139}& \textbf{0.0277}& \textbf{0.0244}& 0.0309& \textbf{0.6487}\\
 \bottomrule
 \end{tabular}
 }
\end{table}

For purposes of this experiment, FOS consistently demonstrates that it improves both accuracy and fairness over baselines. It also outperforms other fairness pre-processing algorithms on a number of measures. Thus, this experiment shows that an oversampling technique that is adopted from imbalanced learning can achieve significant improvements in group fairness measures. This also shows the close relationship between class and protected group imbalance and fairness - by jointly improving class and protected group imbalance ratios, we can affect a substantial improvement in group fairness measures. These results also indicate that it is possible to increase both balanced accuracy and fairness simultaneously (\textbf{RQ1 answered}).

\subsection{Experiment 2: Oversampling for fairness-aware classifiers}

In addition, FOS regularly improves the accuracy and fairness metrics of fairness-aware algorithms (see Table~\ref{tab:fclass}). On the Compas dataset, it uniformly improves results, and in cases where it does not (e.g., equal odds), it misses first place by .0013. In the case of Adult Census, it generally displays better results, and in cases where there is a slight degradation in fairness, there is a substantial improvement in accuracy. Thus, FOS can also supplement existing fairness-aware algorithms as a data pre-processing step (\textbf{RQ2 answered}).

\begin{table*}
\footnotesize
 \caption{Fairness Classifiers With \& Without FOS}
 \label{tab:fclass}
 \scalebox{0.8}{
 \begin{tabular}{lcccccc}
 \toprule
 \textbf{Method}&\textbf{Bal Acc} &\textbf{Avg Odds} & \textbf{Absol Avg Odds}&\textbf{ TNRD}&\textbf{Equal Odds}&\textbf{Fair Utility}\\
 \midrule
 \textbf{German Dataset}&&&&&&\\
 \texttt{Exp. Grad.} & 0.6562& 0.0716& 0.0857& 0.0557& 0.1156& 0.5993 \\
 \texttt{Exp. Grad. (+FOS)}& \textbf{0.7160} & \textbf{0.0306}& \textbf{0.0437}& \textbf{0.0271}& \textbf{0.0602}& \textbf{0.6845}\\
 \midrule
 \texttt{Adver. Debias.}& 0.5739& 0.6006& 0.6006& 0.6143& 0.5868& 0.2613\\
 \texttt{Adver. Debias. (+FOS)}& \textbf{0.6680}& \textbf{0.3988}& \textbf{0.3988}& \textbf{0.3989}& \textbf{0.3986}& \textbf{0.4272}\\
 \midrule
 \texttt{SSR}& 0.6999& \textbf{0.0032}& \textbf{0.0545}& 0.0577& \textbf{0.0513}& \textbf{0.6617}\\
 \texttt{SSR (+FOS)}& \textbf{0.7026}& 0.0677& 0.0720& \textbf{0.0459}& 0.0980& 0.6528\\
 \midrule
 \textbf{Adult Dataset}&&&&&&\\
 \texttt{Exp. Grad.} & 0.6676& \textbf{0.0089}& \textbf{0.0089}& \textbf{0.0114}& \textbf{0.0063}& 0.6617 \\
 \texttt{Exp. Grad. (+FOS)}& \textbf{0.7387} & 0.0107& 0.0133& 0.0154& 0.0112& \textbf{0.7288}\\
 \midrule
 \texttt{Adver. Debias.}& 0.7466& 0.0528& \textbf{0.0594}& 0.0662& \textbf{0.0526}& 0.7022\\
 \texttt{Adver. Debias. (+FOS)}& \textbf{0.8132}& \textbf{0.0430}& 0.0671& \textbf{0.0241}& 0.1101& \textbf{0.7586}\\
 \midrule
 \texttt{SSR}& 0.6976& 0.1047& 0.1047& 0.1327& 0.0768& 0.6246\\
 \texttt{SSR (+FOS)}& \textbf{0.7132}& \textbf{0.0293}& \textbf{0.0325}& \textbf{0.0500}& \textbf{0.0151}& \textbf{0.6901}\\
 \midrule
 \textbf{Compas Dataset}&&&&&&\\
 \texttt{Exp. Grad.} & 0.6638& 0.0167& 0.0327& 0.0379& \textbf{0.0276}& 0.6419 \\
 \texttt{Exp. Grad. (+FOS)}& \textbf{0.6697} & \textbf{0.0147}& \textbf{0.0300}& \textbf{0.0308}& 0.0293& \textbf{0.6492}\\
 \midrule
 \texttt{Adver. Debias.}& 0.6718& 0.2066& 0.2006& 0.2291& 0.1720& 0.5369\\
 \texttt{Adver. Debias. (+FOS)}& \textbf{0.6775}& \textbf{0.1267}& \textbf{0.1267}& \textbf{0.1345}& \textbf{0.1190}& \textbf{0.5922}\\
 \midrule
 \texttt{SSR}& \textbf{0.6251}& 0.0649& 0.0649& 0.0750& 0.0549& 0.5848\\
 \texttt{SSR (+FOS)}& 0.6521& \textbf{0.0142}& \textbf{0.0184}& \textbf{0.0161}& \textbf{0.0207}& \textbf{0.6440}\\
\bottomrule
 \end{tabular}
 }
\end{table*}

\subsection{Experiment 3: Robustness to increasing imbalance ratios}

\begin{figure}
 \centering
 \begin{subfigure}[b]{0.3\textwidth}
 \centering
 \includegraphics[width=\textwidth]{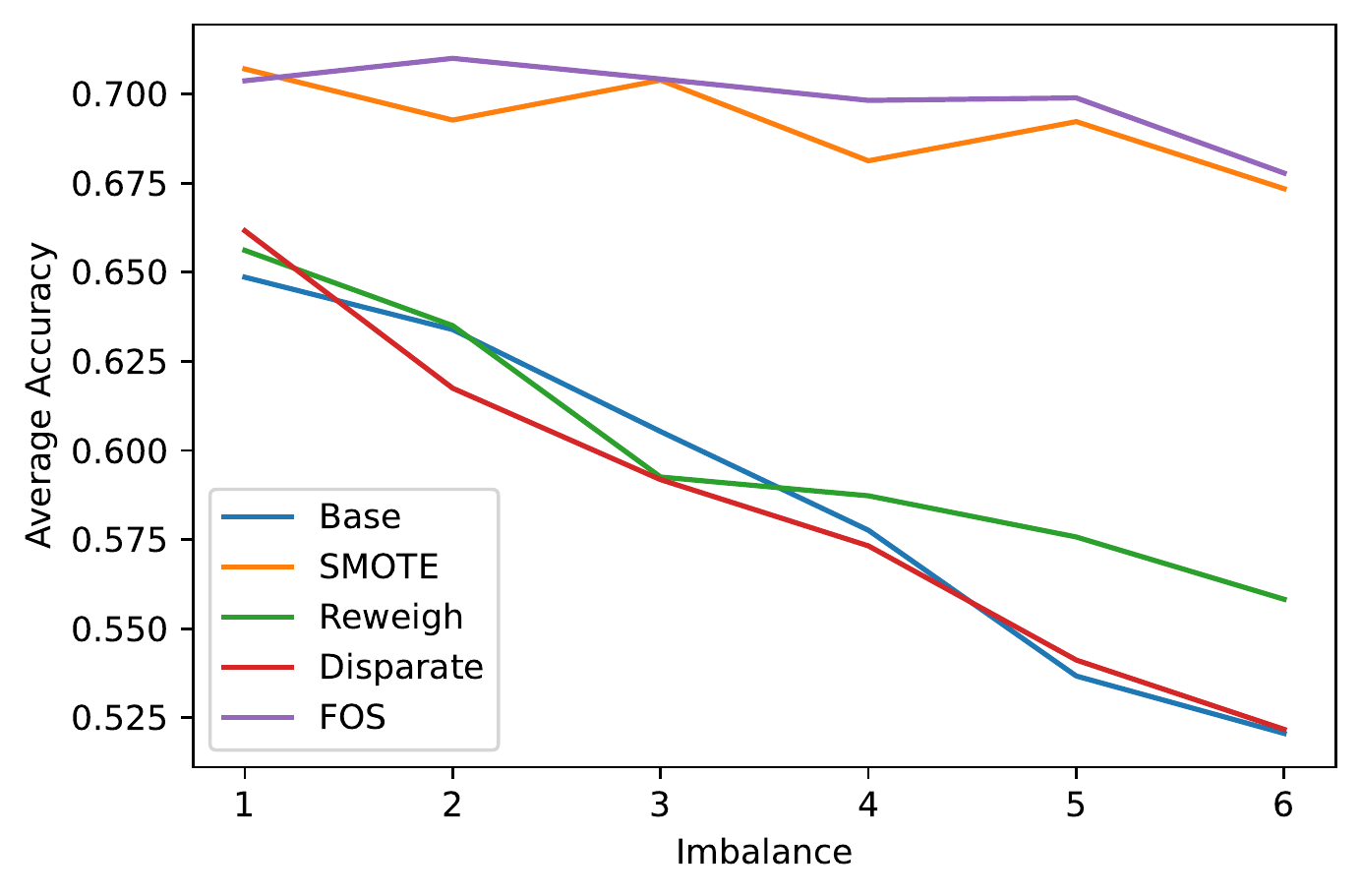}
 \caption{German Balanced Accuracy}
 \label{fig:fig6}
 \end{subfigure}
 \hfill
 \begin{subfigure}[b]{0.3\textwidth}
 \centering
 \includegraphics[width=\textwidth]{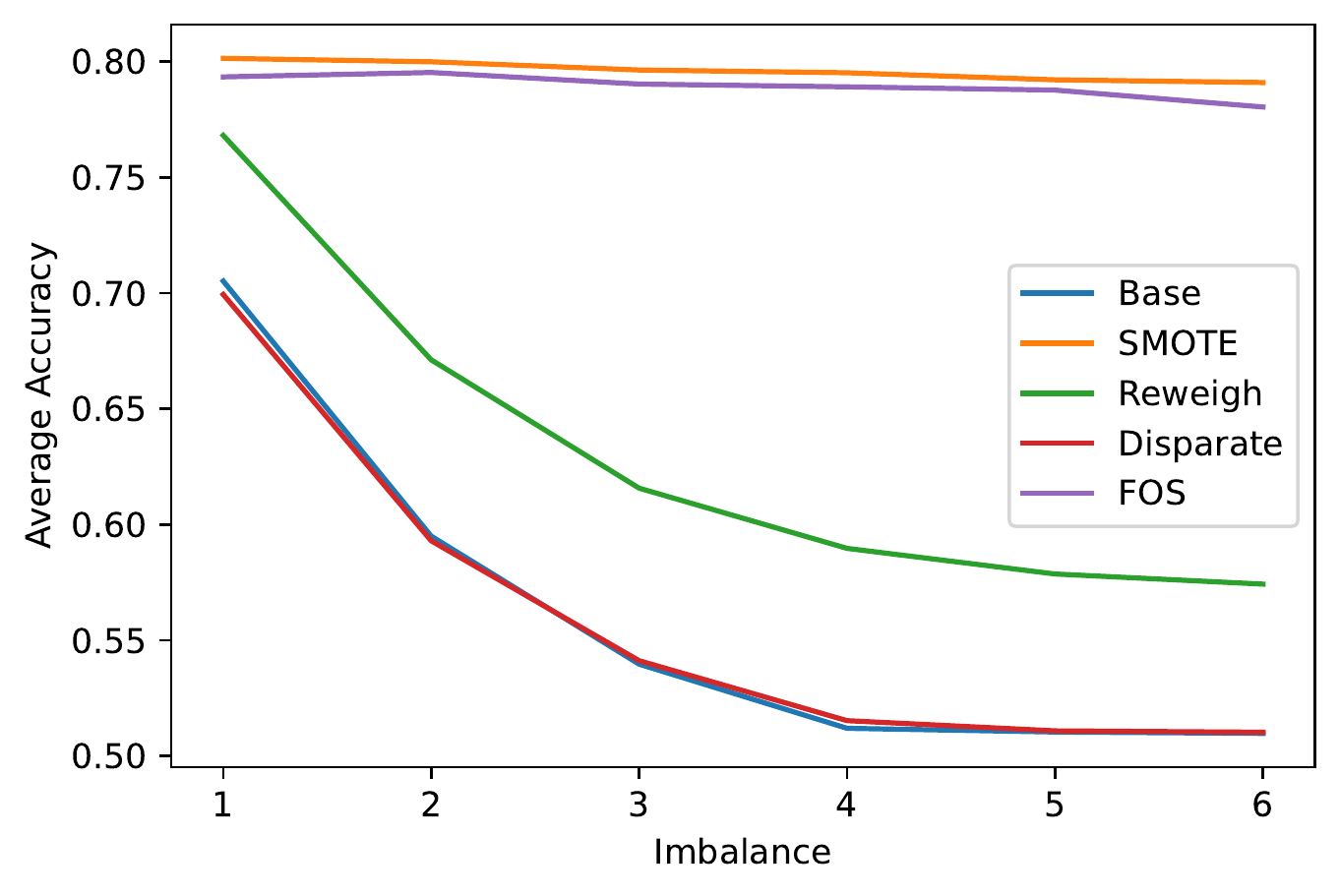}
 \caption{Adult Balanced Accuracy}
 \label{fig:fig7}
 \end{subfigure}
 \hfill
 \begin{subfigure}[b]{0.3\textwidth}
 \centering
 \includegraphics[width=\textwidth]{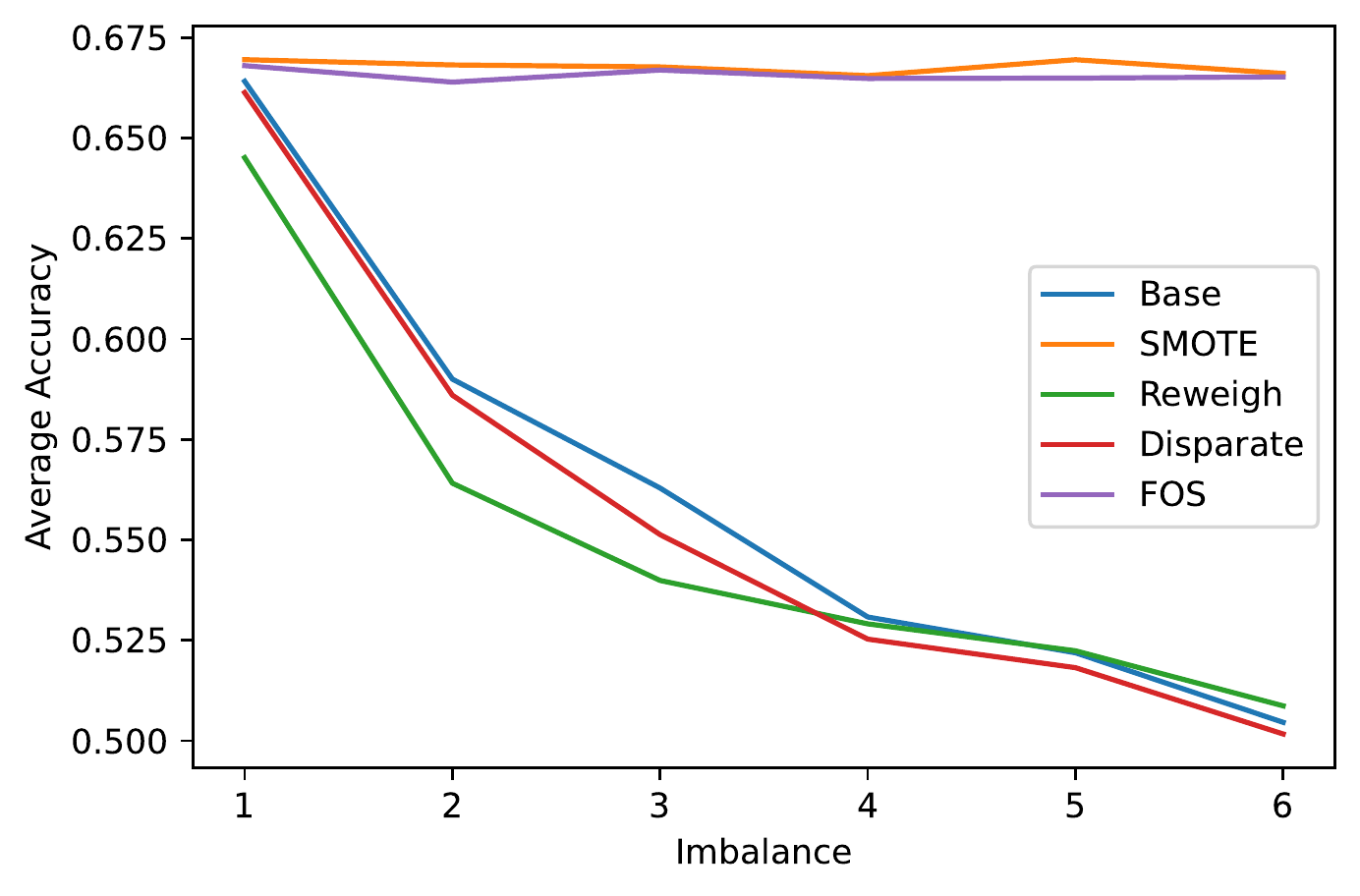}
 \caption{Compas Balanced Accuracy}
 \label{fig:fig8}
 \end{subfigure}
 
 \begin{subfigure}[b]{0.3\textwidth}
 \centering
 \includegraphics[width=\textwidth]{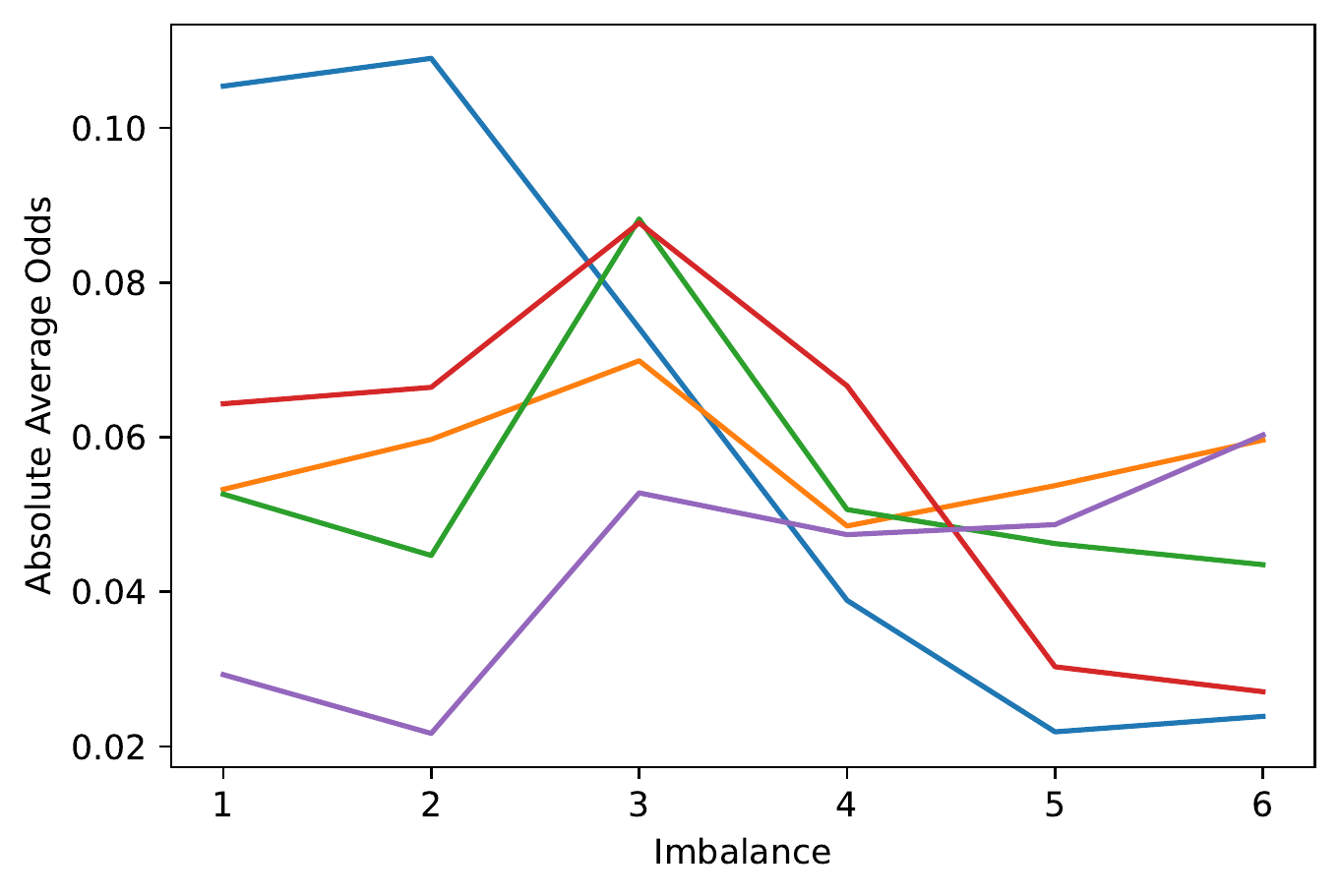}
 \caption{German Credit AAO}
 \label{fig:fig9}
 \end{subfigure}
 \hfill
 \begin{subfigure}[b]{0.3\textwidth}
 \centering
 \includegraphics[width=\textwidth]{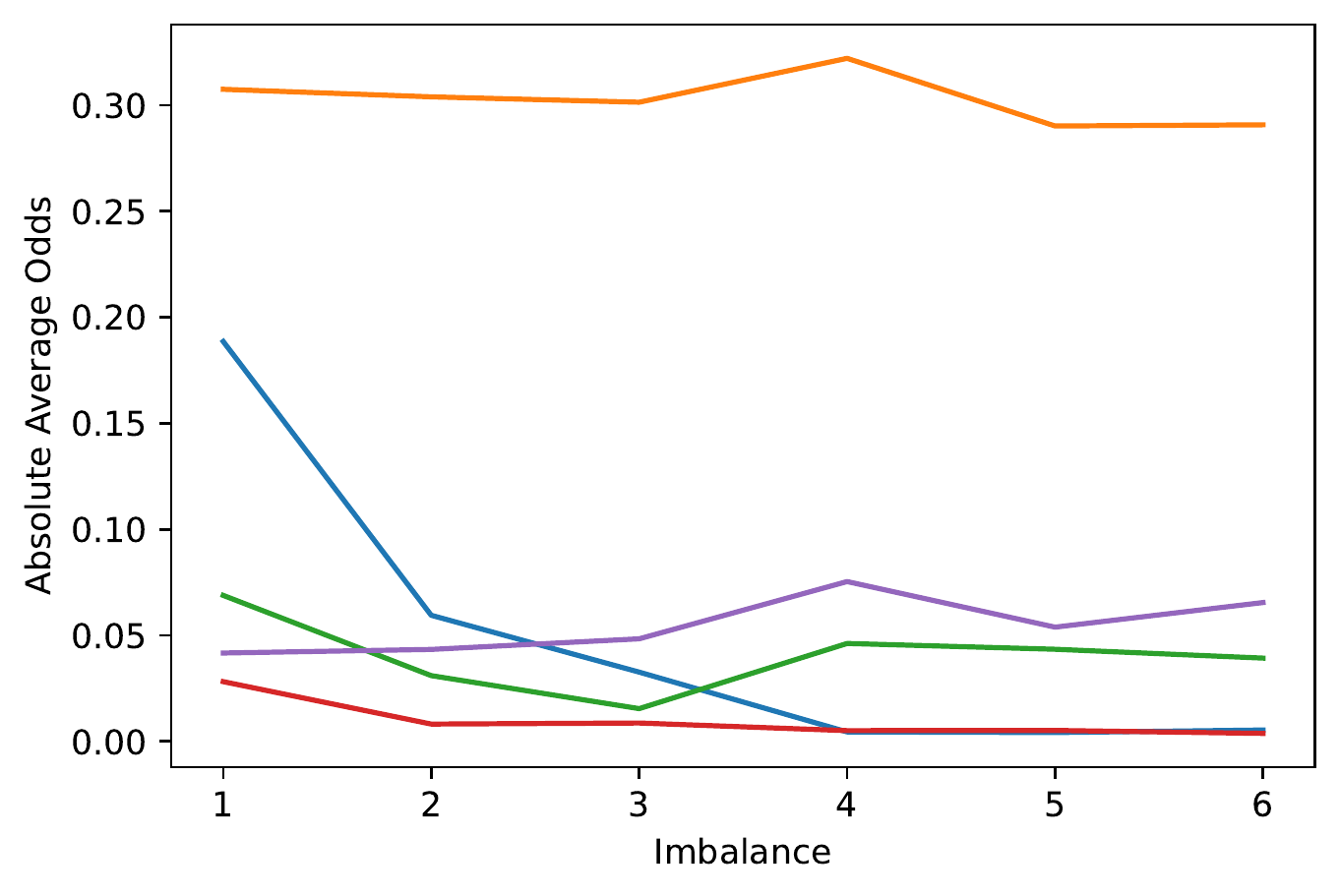}
 \caption{Adult Census AAO}
 \label{fig:fig10}
 \end{subfigure}
 \hfill
 \begin{subfigure}[b]{0.3\textwidth}
 \centering
 \includegraphics[width=\textwidth]{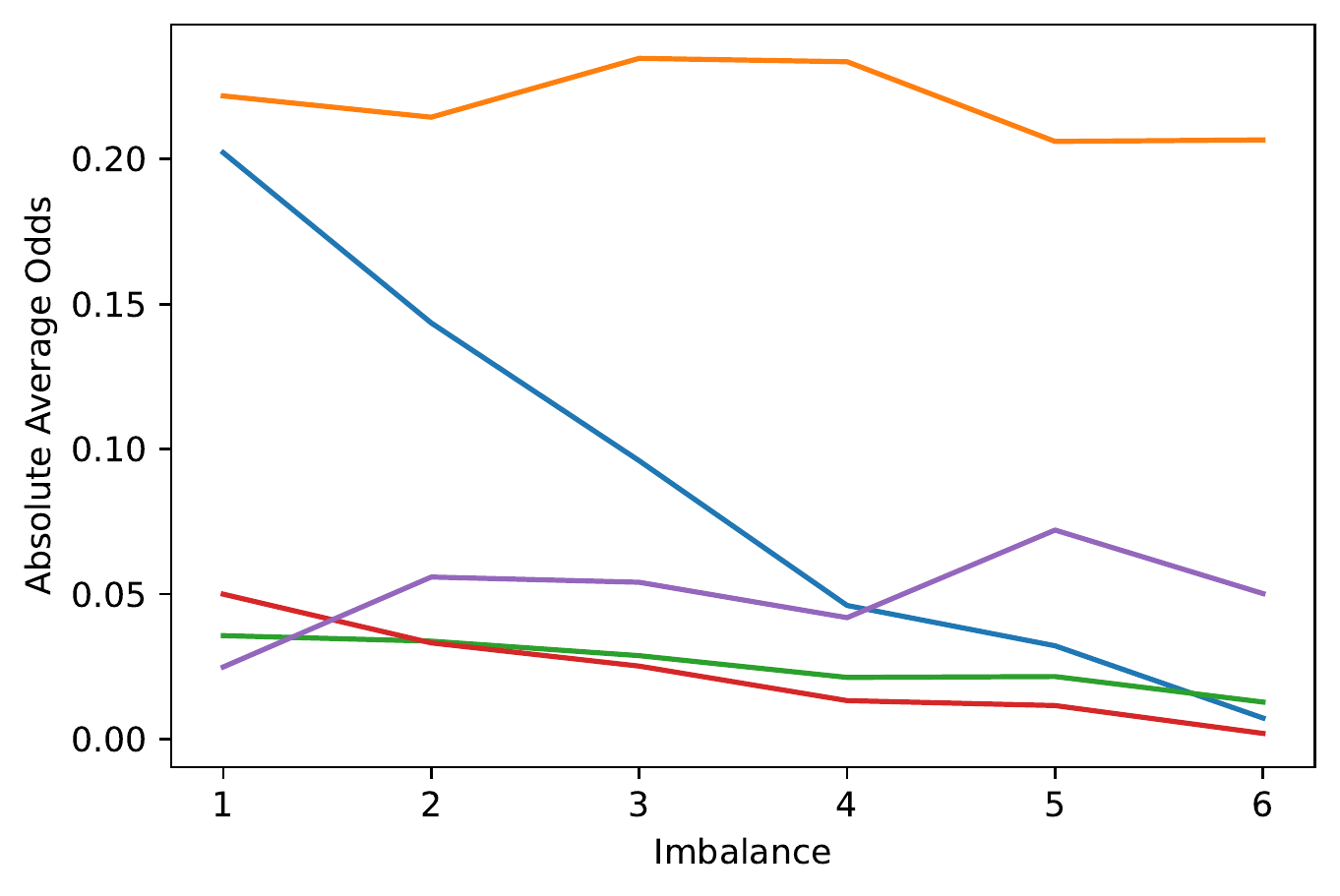}
 \caption{Compas AAO}
 \label{fig:fig11}
 \end{subfigure}
 
 \begin{subfigure}[b]{0.3\textwidth}
 \centering
 \includegraphics[width=\textwidth]{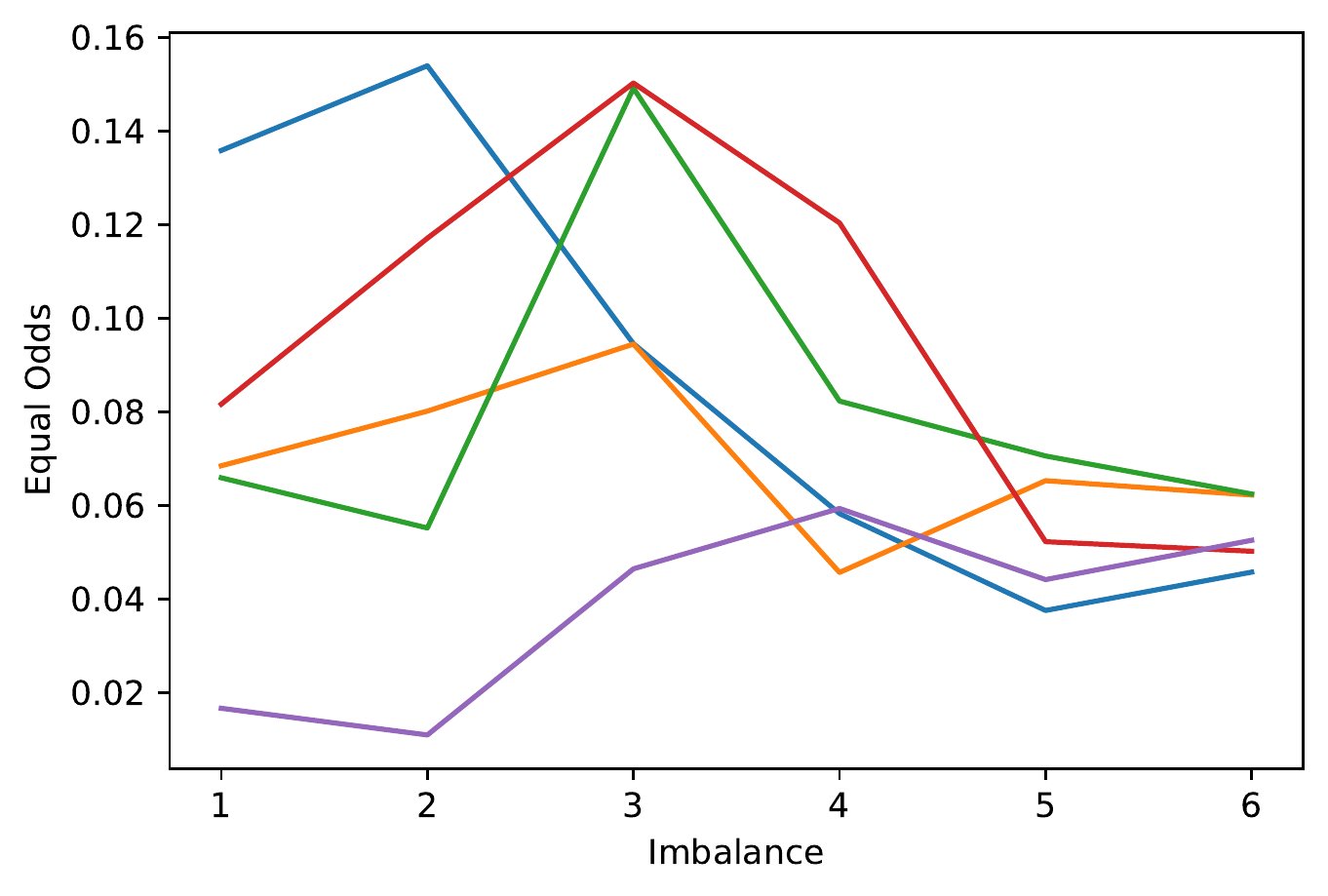}
 \caption{German EO}
 \label{fig:fig12}
 \end{subfigure}
 \hfill
 \begin{subfigure}[b]{0.3\textwidth}
 \centering
 \includegraphics[width=\textwidth]{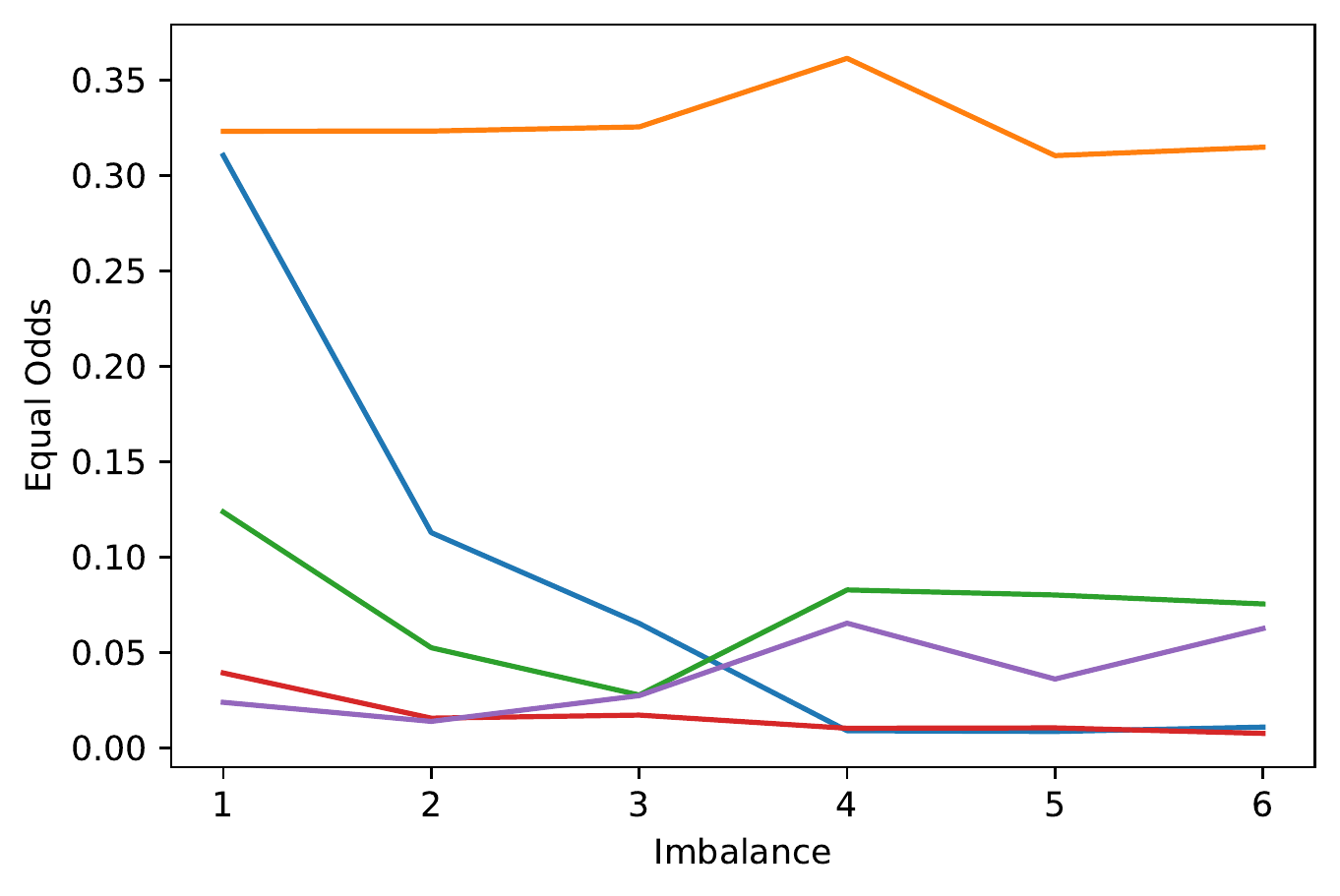}
 \caption{Adult EO}
 \label{fig:fig13}
 \end{subfigure}
 \hfill
 \begin{subfigure}[b]{0.3\textwidth}
 \centering
 \includegraphics[width=\textwidth]{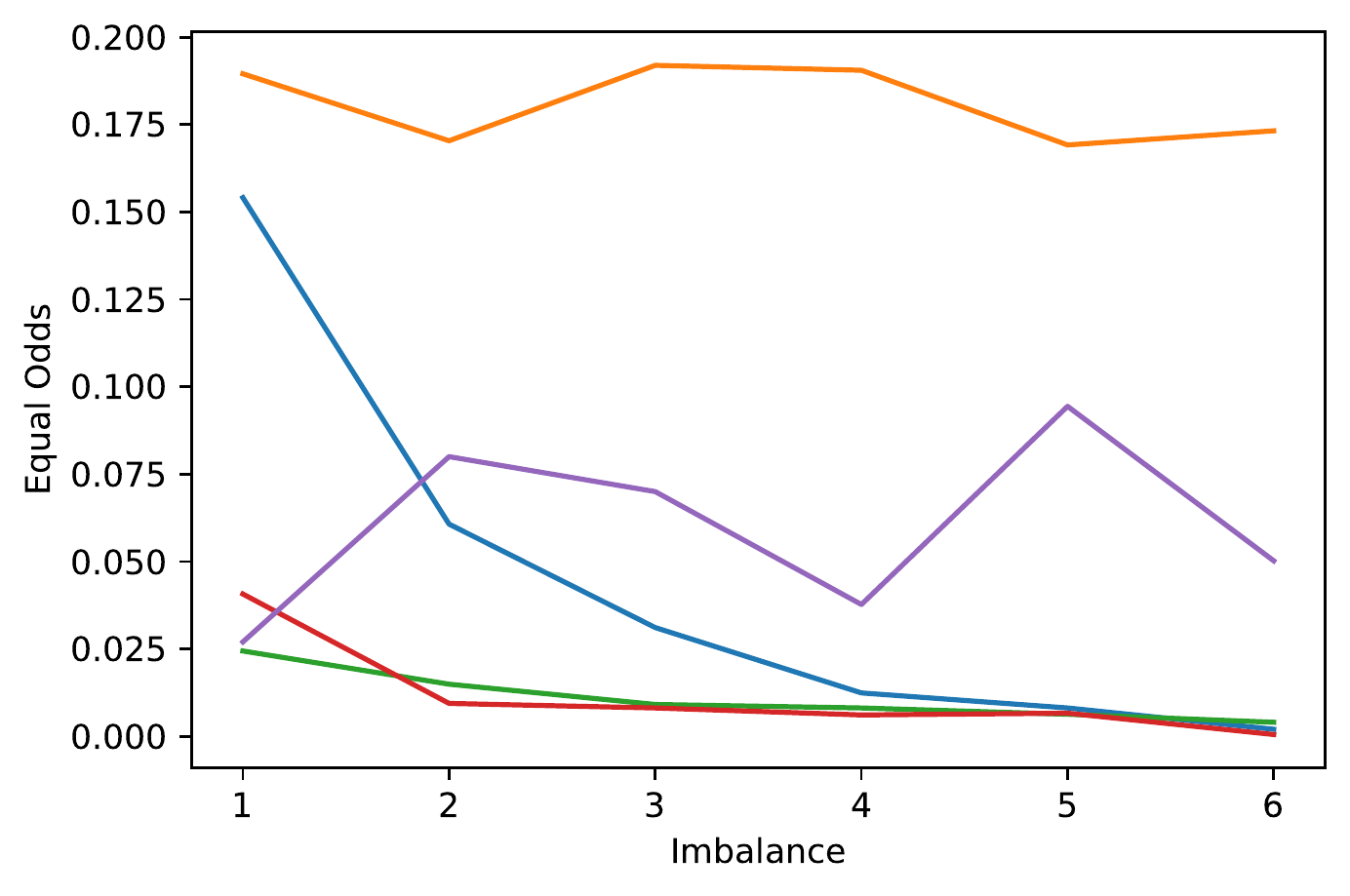}
 \caption{Compas EO}
 \label{fig:fig14}
 \end{subfigure}
 
 \begin{subfigure}[b]{0.3\textwidth}
 \centering
 \includegraphics[width=\textwidth]{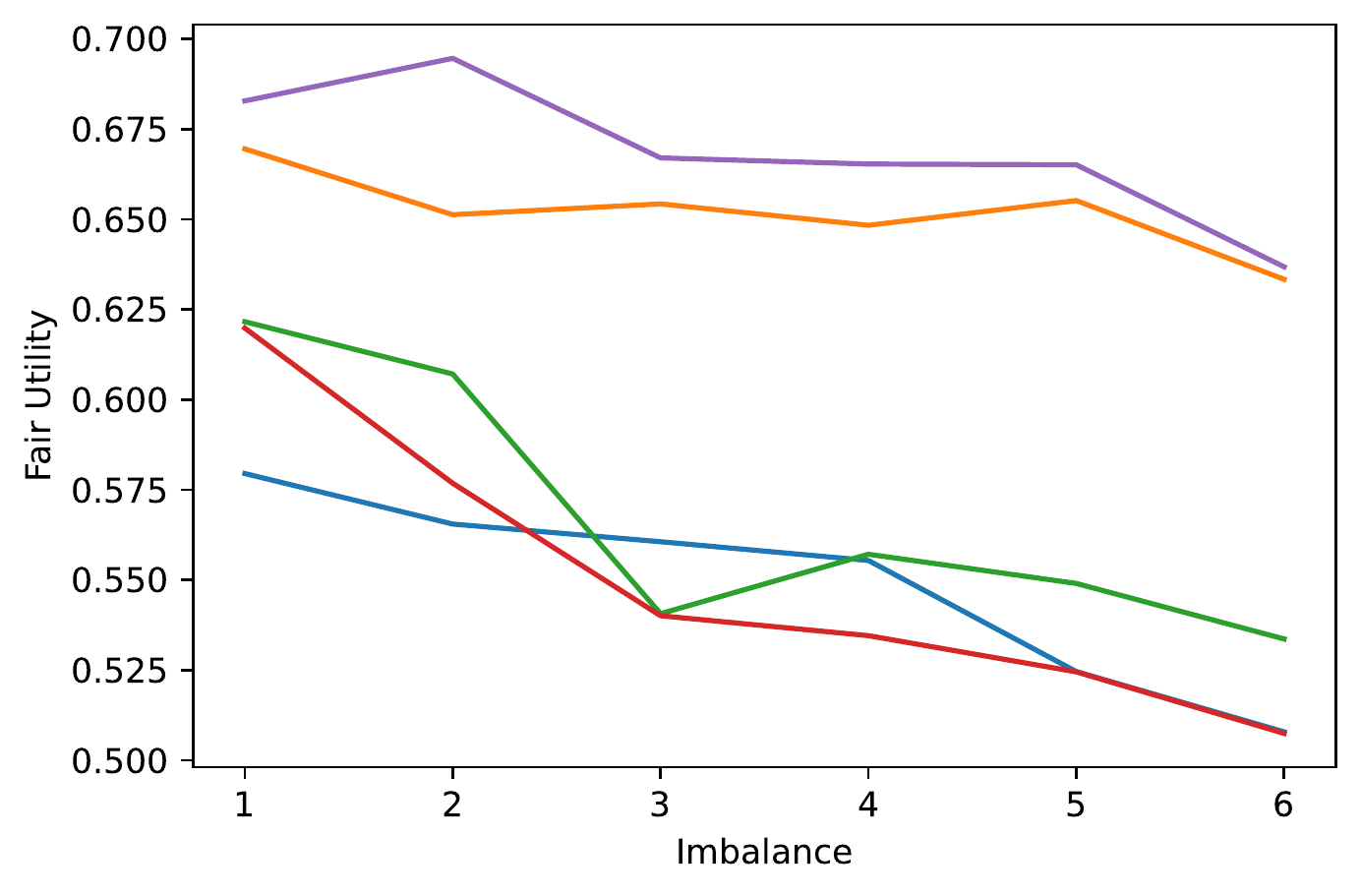}
 \caption{German Fair Utility}
 \label{fig:fig15}
 \end{subfigure}
 \hfill
 \begin{subfigure}[b]{0.3\textwidth}
 \centering
 \includegraphics[width=\textwidth]{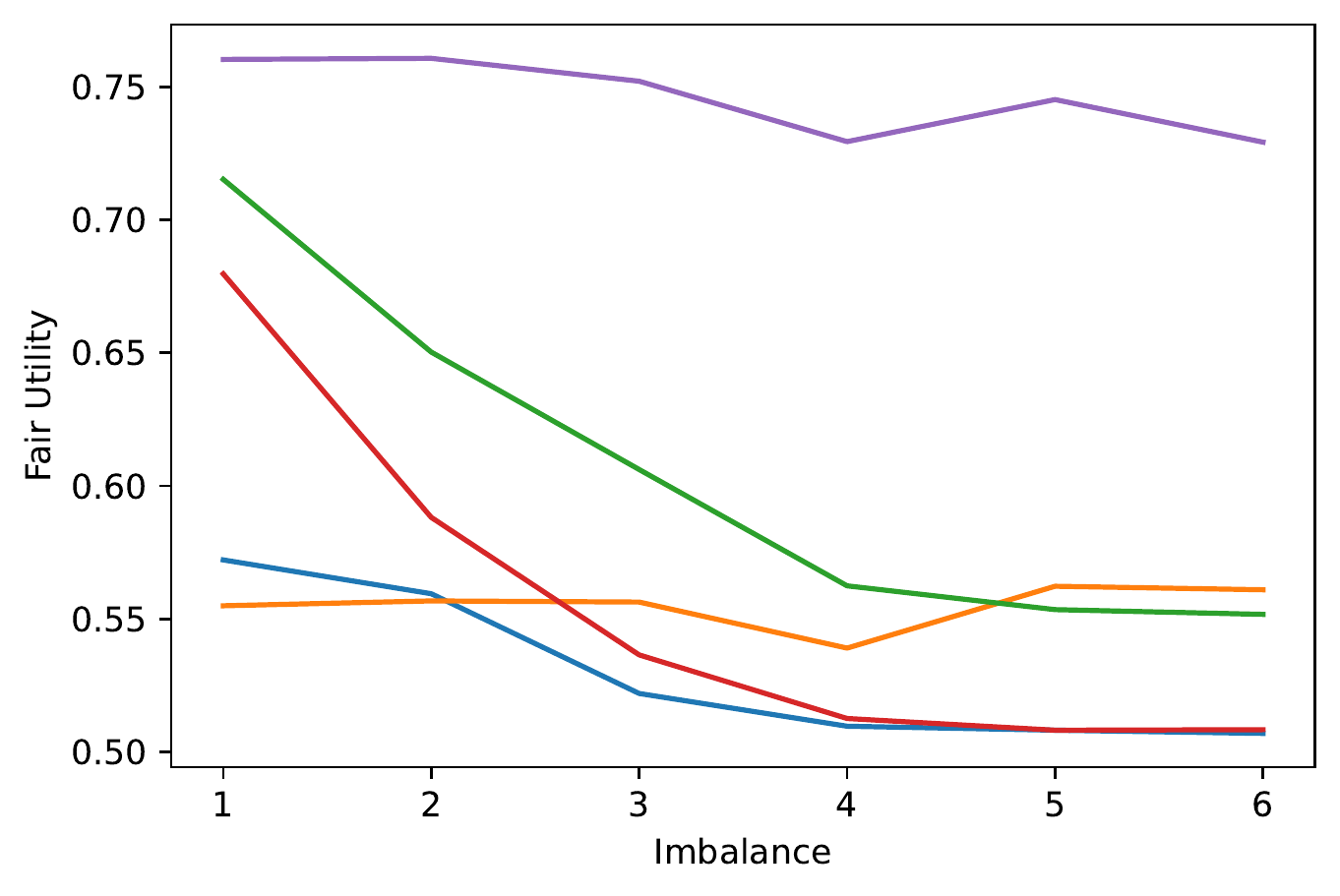}
 \caption{Adult Fair Utility}
 \label{fig:fig16}
 \end{subfigure}
 \hfill
 \begin{subfigure}[b]{0.3\textwidth}
 \centering
 \includegraphics[width=\textwidth]{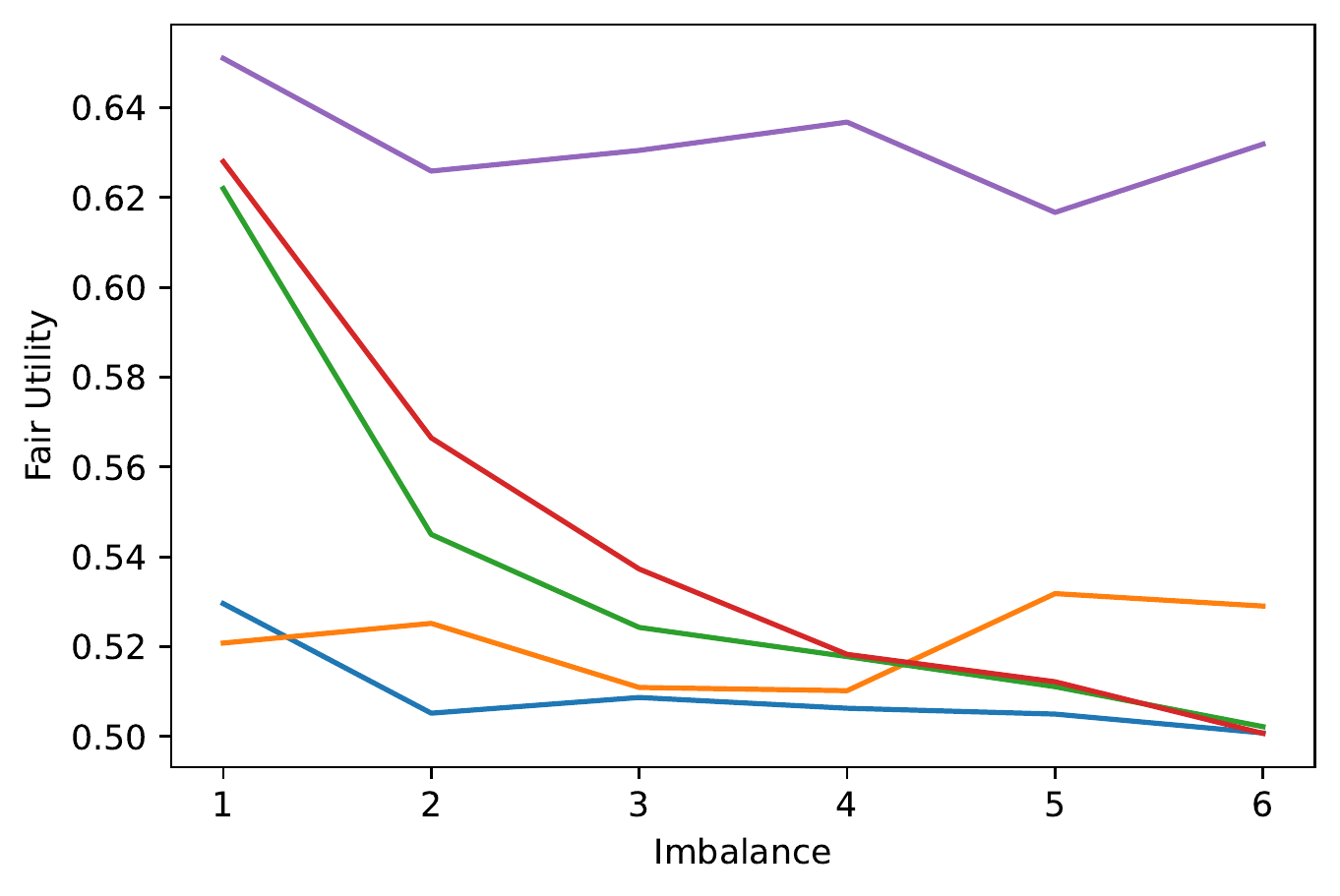}
 \caption{Compas Fair Utility}
 \label{fig:fig17}
 \end{subfigure}
 
 \caption{\textbf{Impact of Varying Imbalance Levels.} This figure illustrates the impact of varying protected group, minority class imbalance ratios on key performance metrics using a SVM classifier. It compares the performance of the baseline (no training set modifications), 3 benchmark algorithms (SMOTE, Reweighing, and Disparate Impact, and FOS. FOS shows high resilience to increasing imbalance levels with respect to Balanced Accuracy and Fair Utility; while exhibiting less stability with respect to AAO and EO. We hypothesize that the reason for this seeming instability is that the baseline and other benchmarks focus (or favor) majority class predictions over minority class predictions; whereas FOS maintains an even balance between majority and minority class predictions. We can see this in the next figure.}
 \label{fig:ImbalLevels}
\end{figure}

\begin{figure}
 \centering
 \begin{subfigure}[b]{0.3\textwidth}
 \centering
 \includegraphics[width=\textwidth]{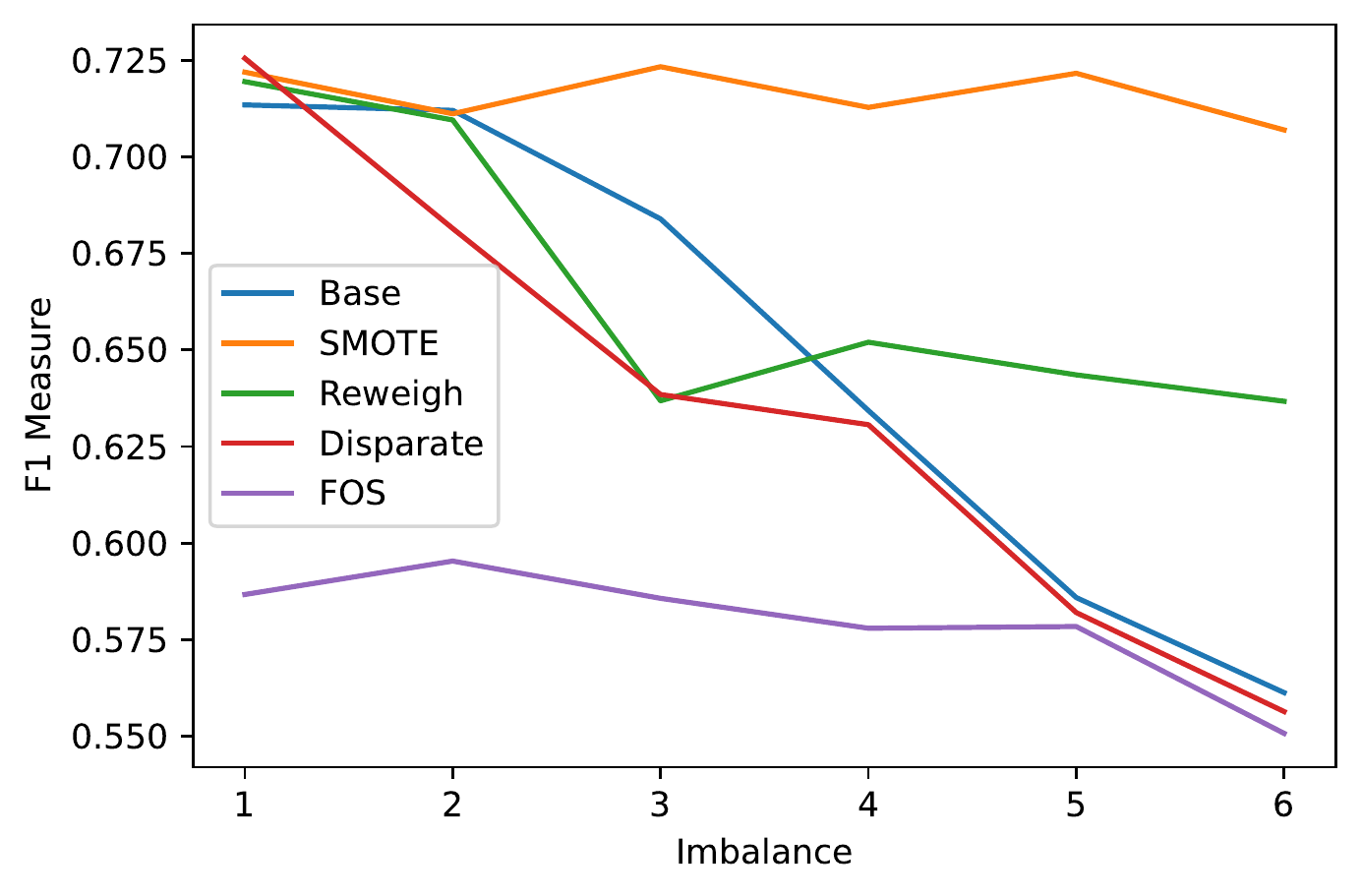}
 \caption{German Credit F1 Measure}
 \label{fig:fig30}
 \end{subfigure}
 \hfill
 \begin{subfigure}[b]{0.3\textwidth}
 \centering
 \includegraphics[width=\textwidth]{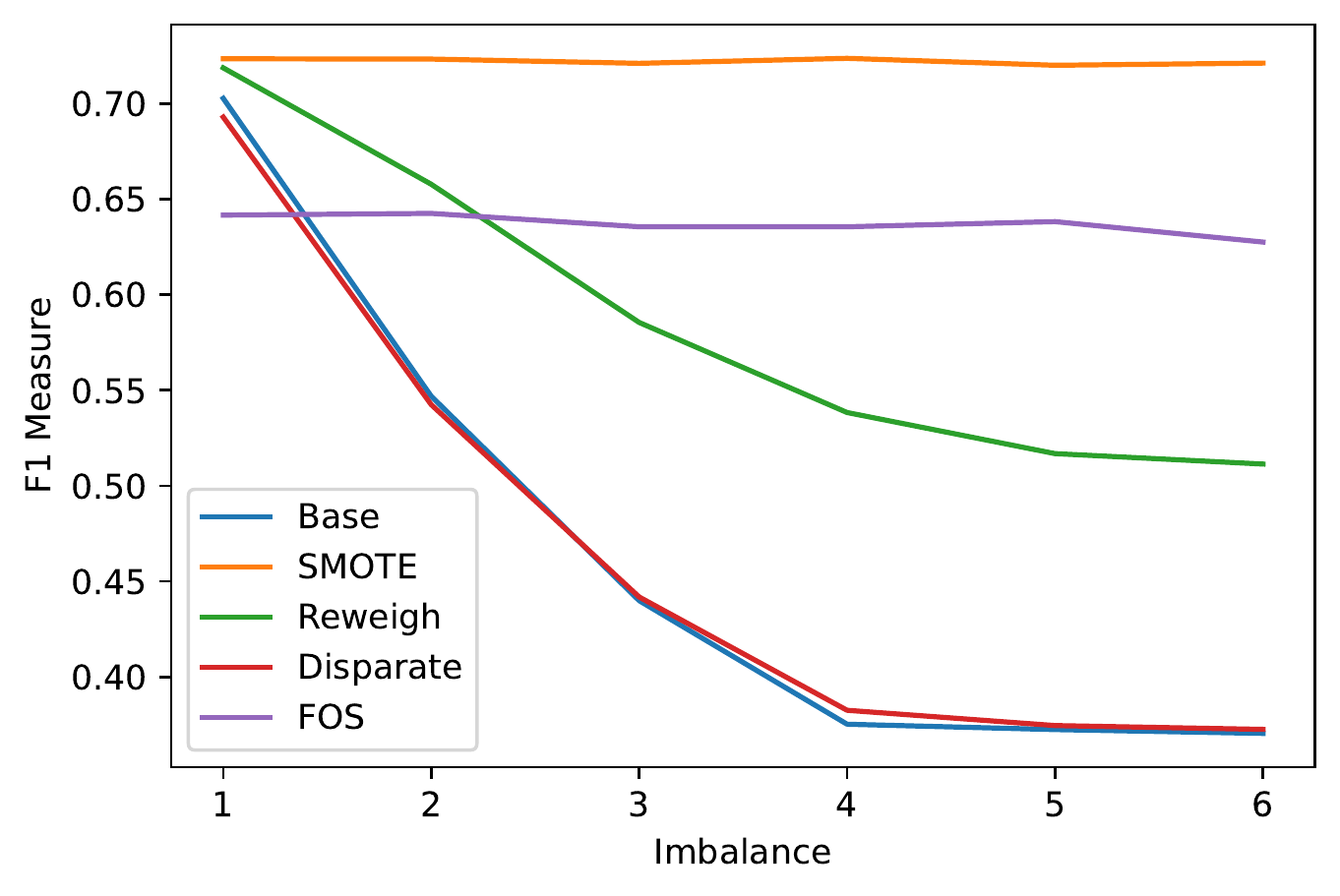}
 \caption{Adult Census F1 Measure}
 \label{fig:fig31}
 \end{subfigure}
 \hfill
 \begin{subfigure}[b]{0.3\textwidth}
 \centering
 \includegraphics[width=\textwidth]{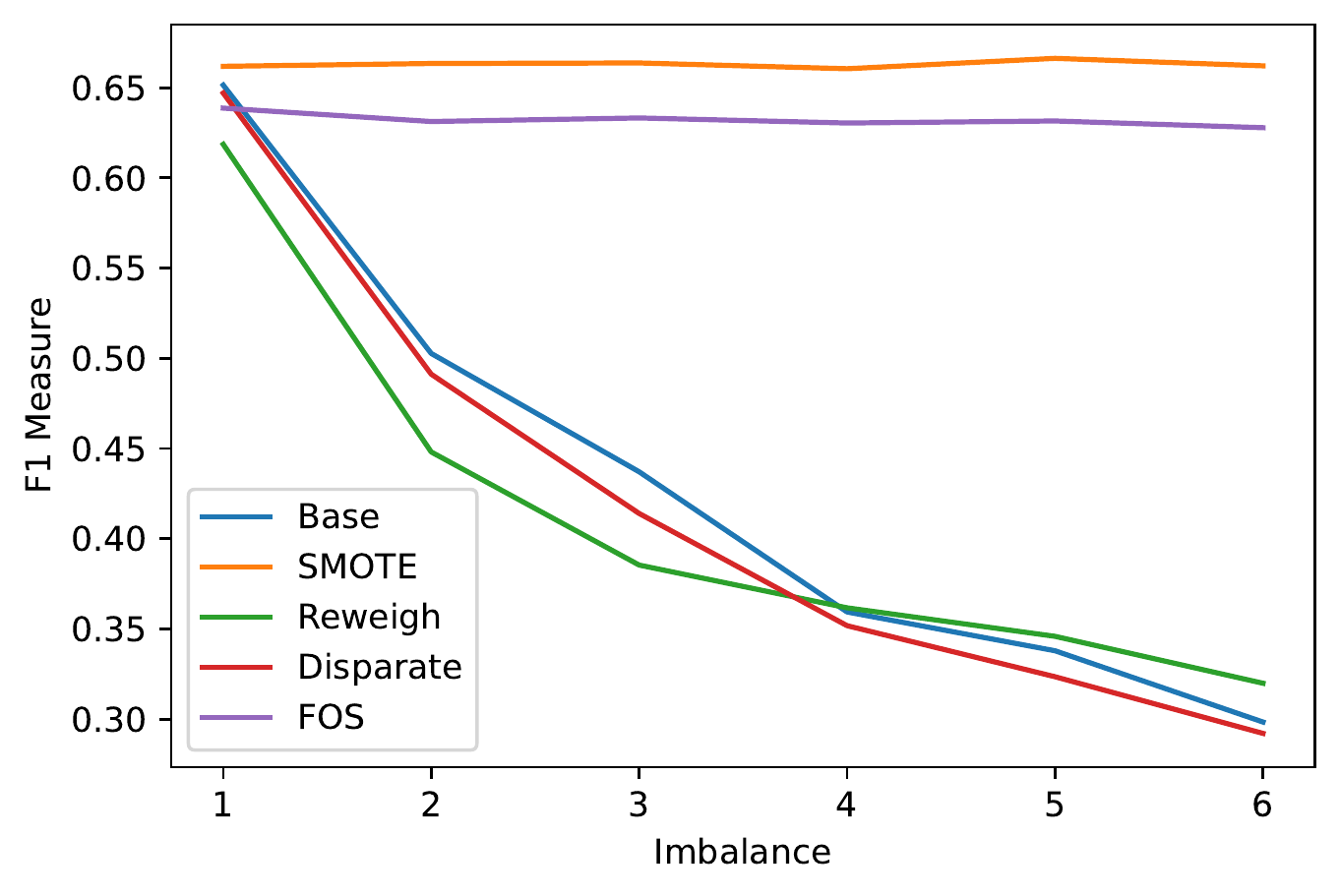}
 \caption{Compas F1 Measure}
 \label{fig:fig32}
 \end{subfigure}
 
 \begin{subfigure}[b]{0.3\textwidth}
 \centering
 \includegraphics[width=\textwidth]{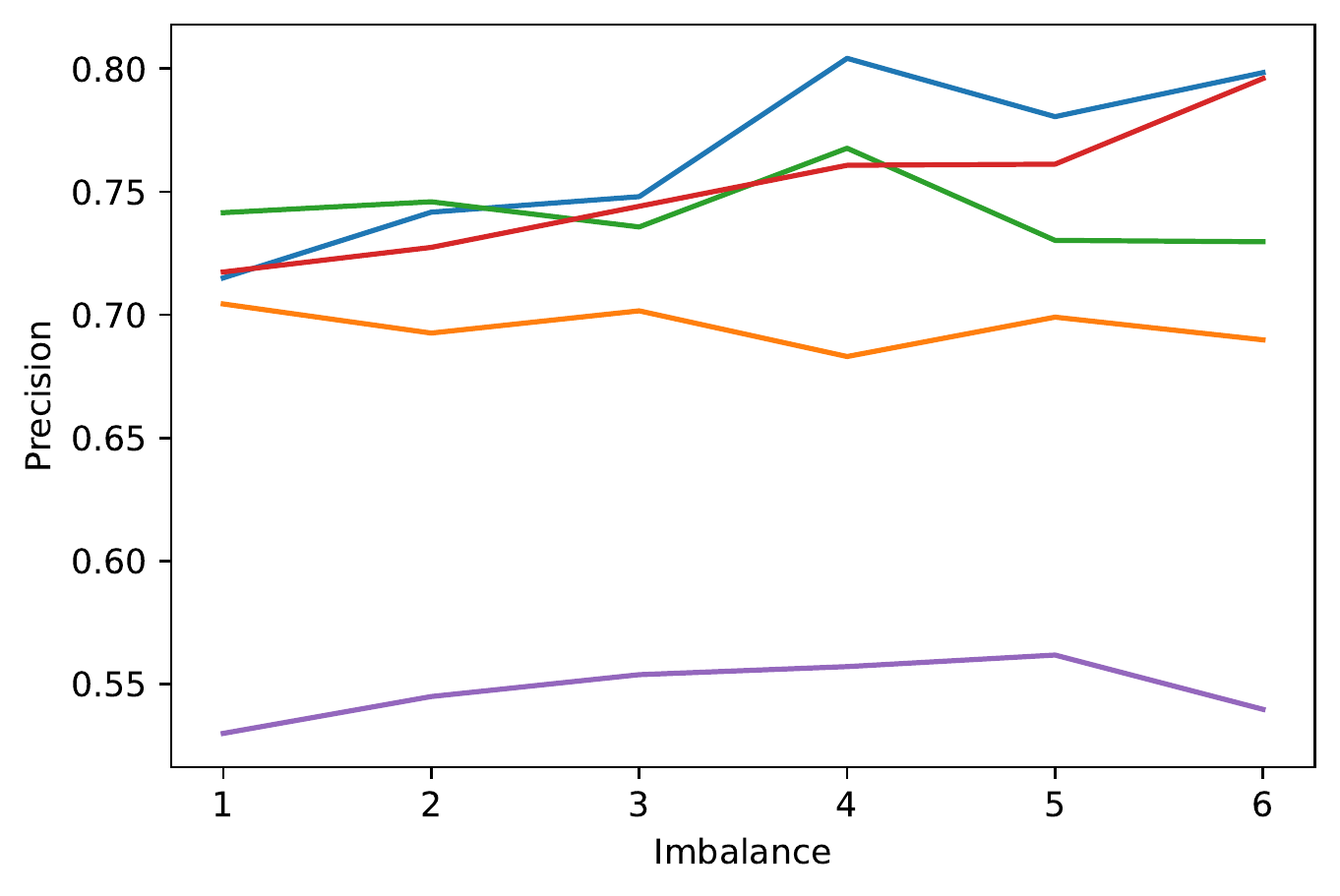}
 \caption{German Credit Precision}
 \label{fig:fig33}
 \end{subfigure}
 \hfill
 \begin{subfigure}[b]{0.3\textwidth}
 \centering
 \includegraphics[width=\textwidth]{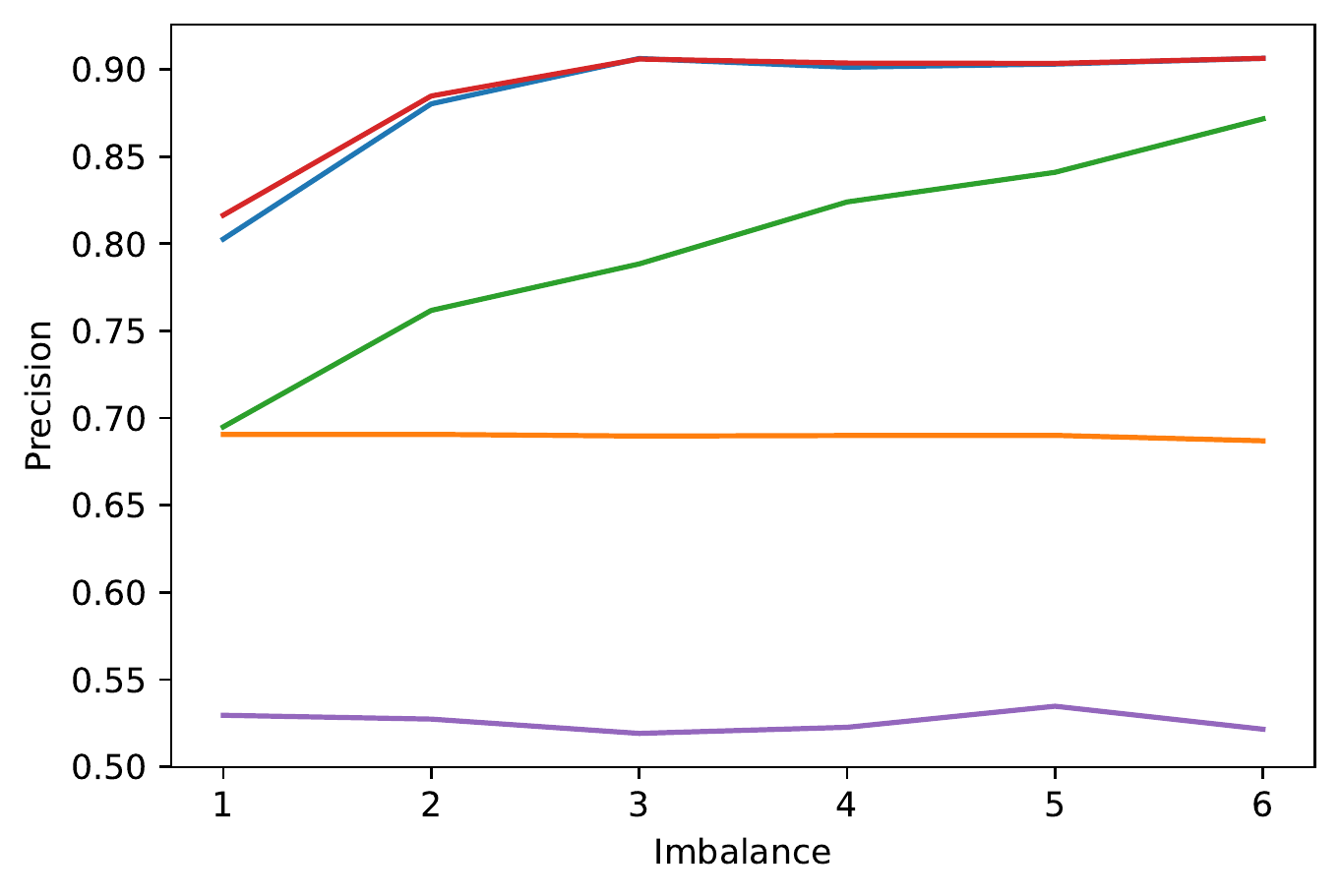}
 \caption{Adult Census Precision}
 \label{fig:fig34}
 \end{subfigure}
 \hfill
 \begin{subfigure}[b]{0.3\textwidth}
 \centering
 \includegraphics[width=\textwidth]{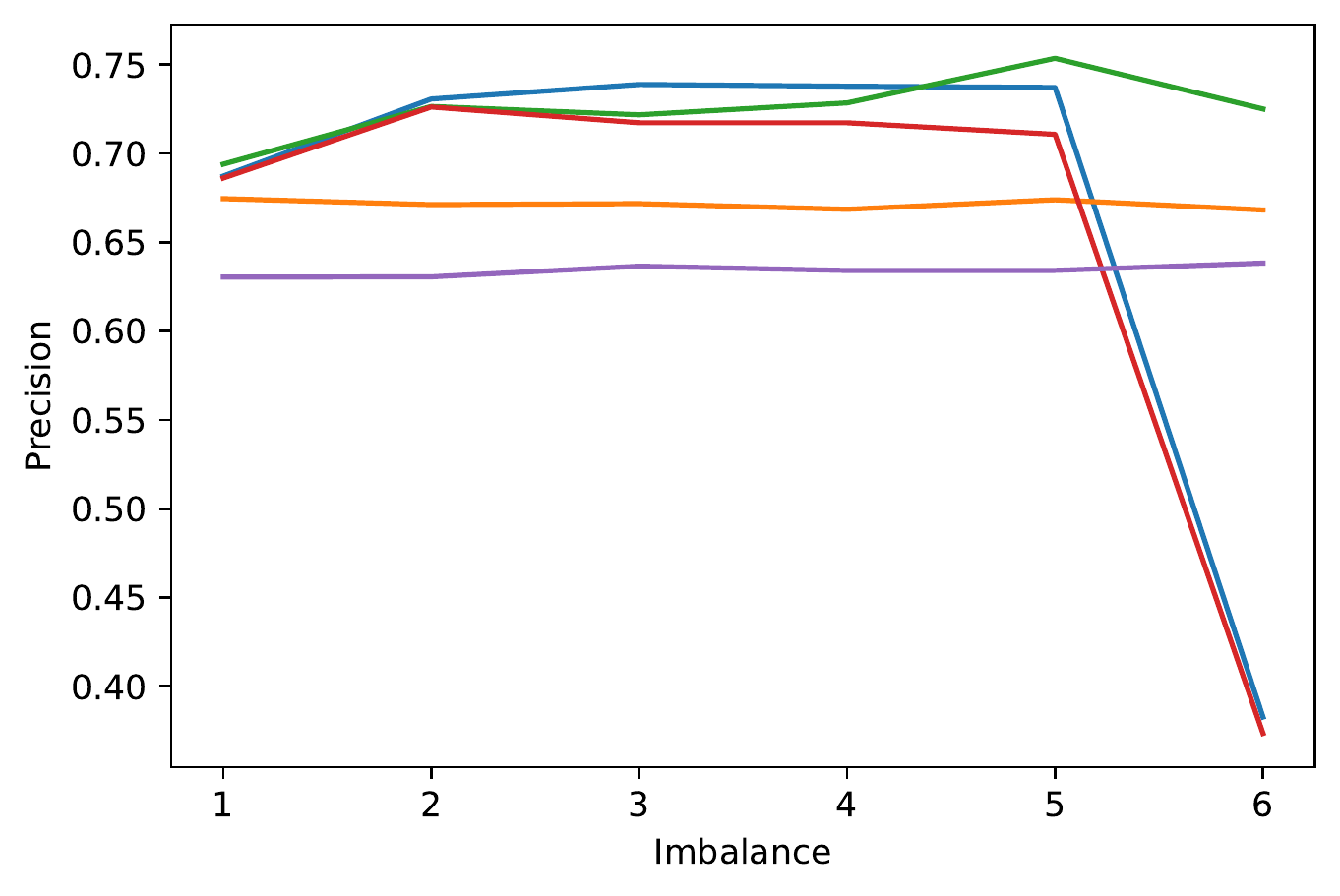}
 \caption{Compas Precision}
 \label{fig:fig35}
 \end{subfigure}
 
 \begin{subfigure}[b]{0.3\textwidth}
 \centering
 \includegraphics[width=\textwidth]{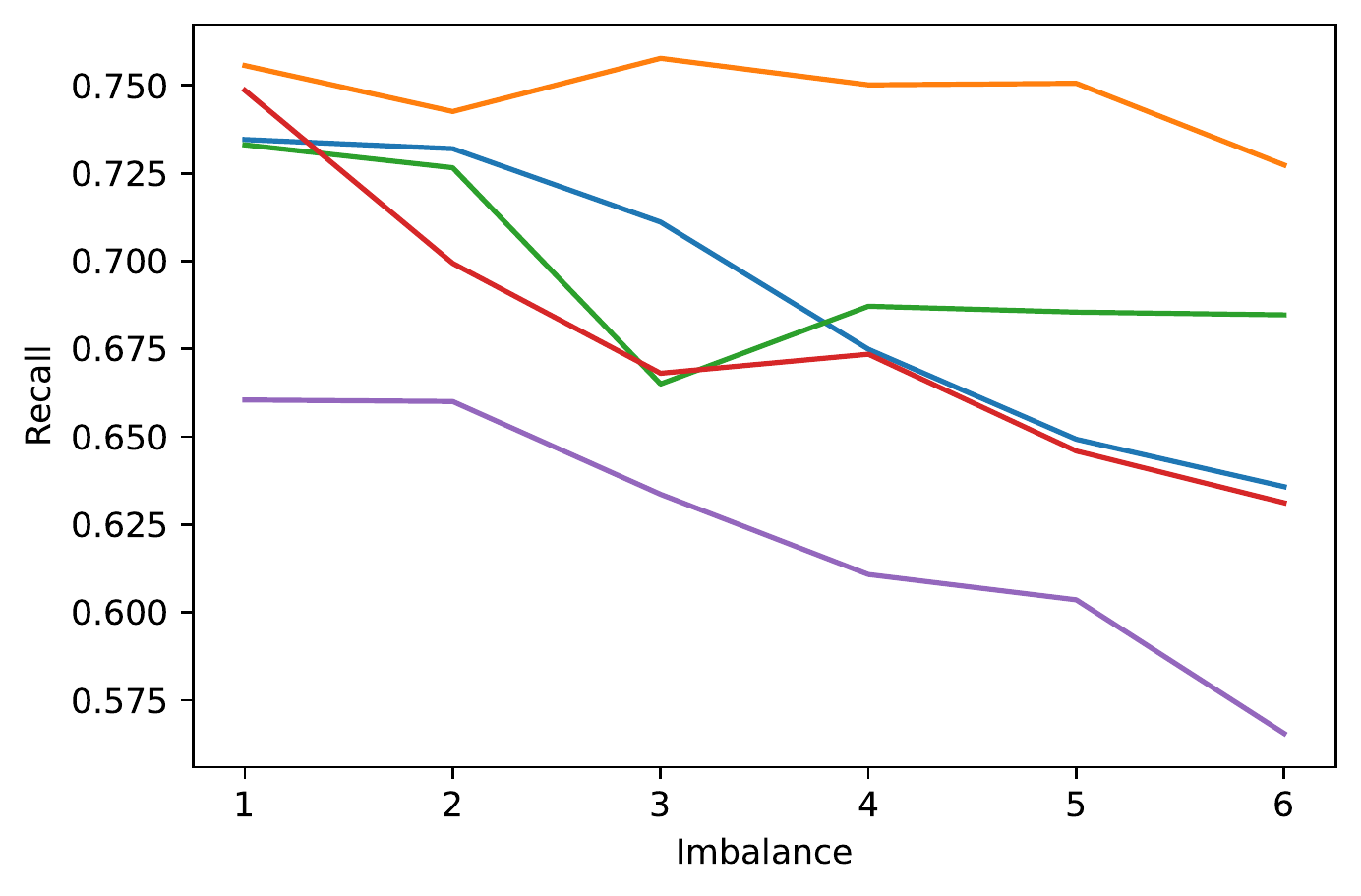}
 \caption{German Credit Recall}
 \label{fig:fig36}
 \end{subfigure}
 \hfill
 \begin{subfigure}[b]{0.3\textwidth}
 \centering
 \includegraphics[width=\textwidth]{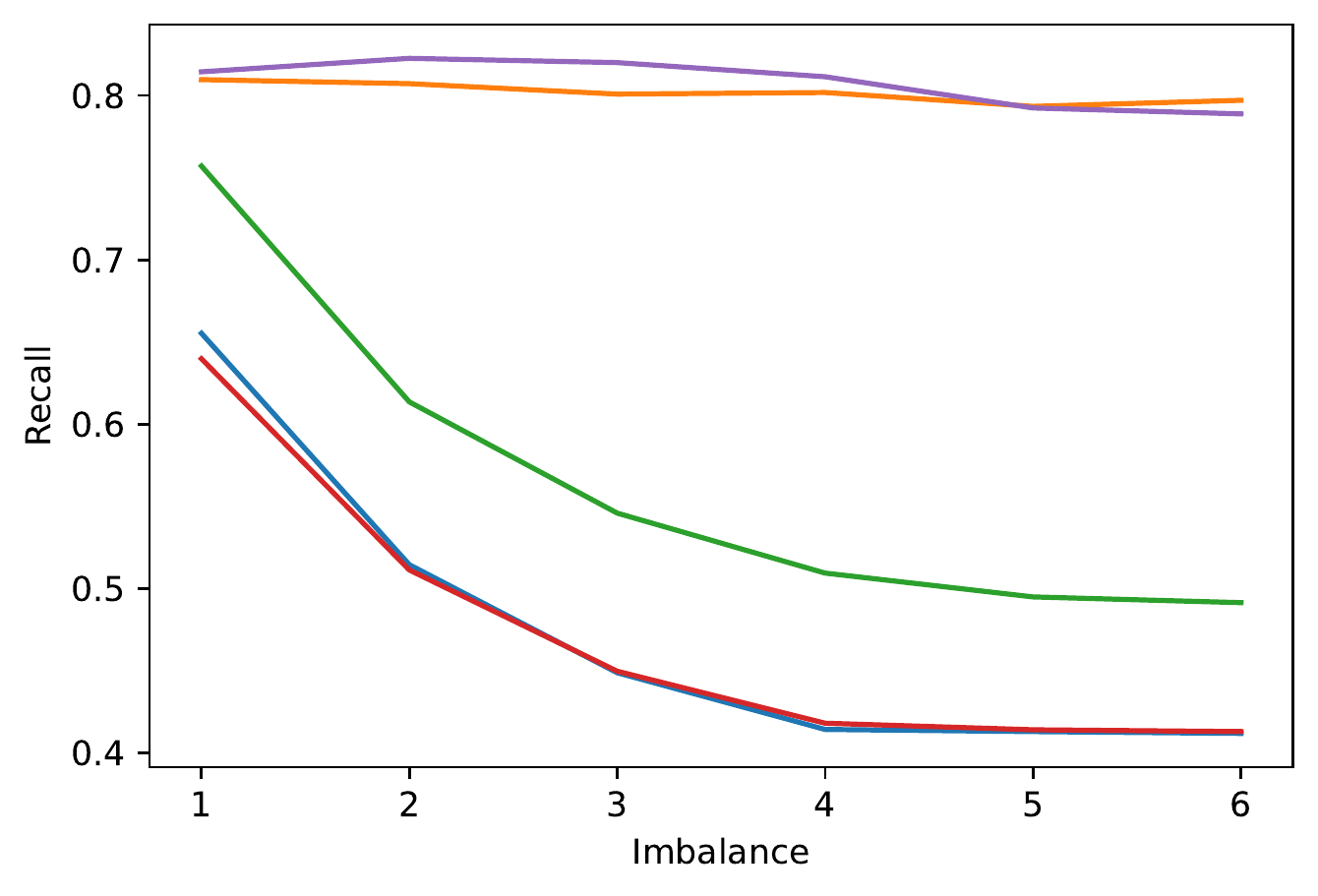}
 \caption{Adult Census Recall}
 \label{fig:fig37}
 \end{subfigure}
 \hfill
 \begin{subfigure}[b]{0.3\textwidth}
 \centering
 \includegraphics[width=\textwidth]{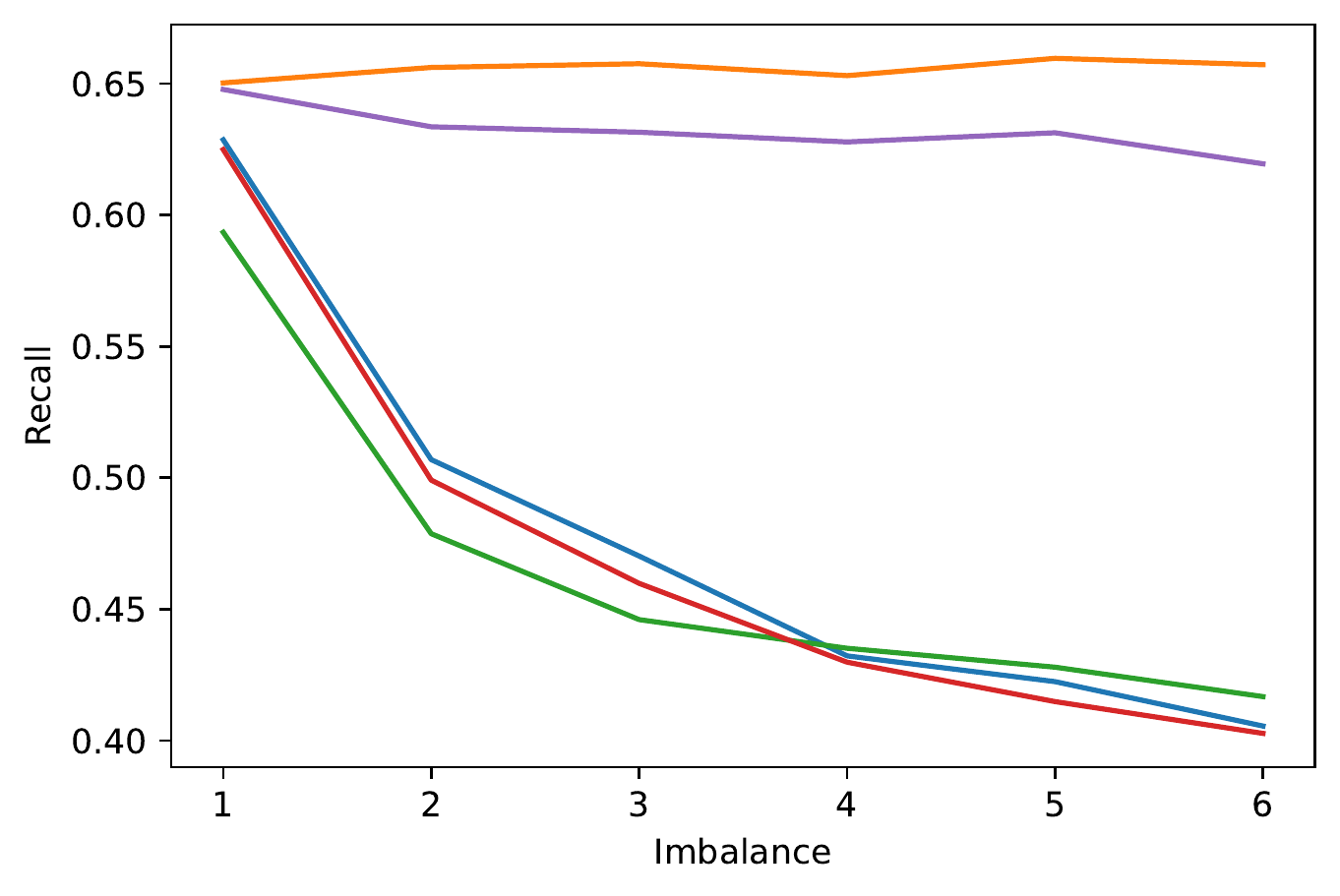}
 \caption{Compas Recall}
 \label{fig:fig38}
 \end{subfigure}
 
 \caption{\textbf{Varying Imbalance Levels on F1, Precision \& Recall.} This figure illustrates the impact of increasing imbalance levels on the F1, precision, and recall measures. For the Compas and Adult Census datasets, which experience relatively more class and protected group imbalance, FOS shows greater resilience.}
 \label{fig:F1}
\end{figure}

FOS performs at the top of the benchmark group in terms of Balanced Accuracy and Fair Utility under increasing levels of imbalance on all three datasets. See Figure~\ref{fig:ImbalLevels}. However, at first glance, it does not outperform in terms of discrimination mitigation at higher levels of imbalance. Upon closer inspection, we believe that the reason why the baseline and other algorithms appear to be more stable at higher imbalance ratios is because their predictions focus on the true positives at the expense of true negatives. This can be seen in the Adult Census and Compas datasets, which have higher levels of imbalance. In those cases, as depicted in Figure~\ref{fig:F1}, the precision ratios increased and the recall ratios decreased for most algorithms, except for FOS and SMOTE (\textbf{RQ3 answered}). As discussed in the Experiments section, it should be remembered that other pre-processing techniques initially failed at imbalance levels greater than 2 and 4 on the German Credit and Compas datasets, respectively. 

\begin{figure*}
 \centering
 \begin{subfigure}[b]{0.25\textwidth}
 \centering
 \includegraphics[width=\textwidth]{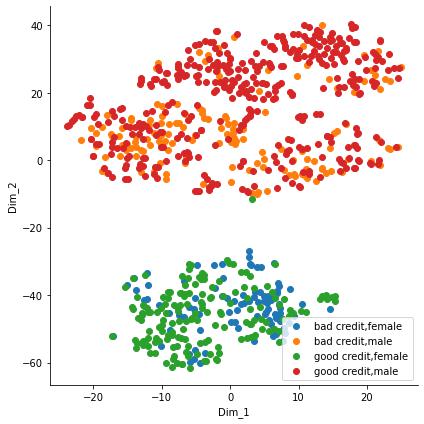}
 \caption{Baseline German Credit data}
 \label{fig:fig1}
 \end{subfigure}
 \hfill
 \begin{subfigure}[b]{0.25\textwidth}
 \centering
 \includegraphics[width=\textwidth]{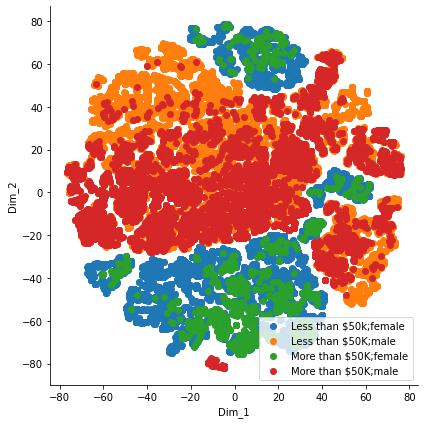}
 \caption{Baseline Adult Census data}
 \label{fig:fig2}
 \end{subfigure}
 \hfill
 \begin{subfigure}[b]{0.25\textwidth}
 \centering
 \includegraphics[width=\textwidth]{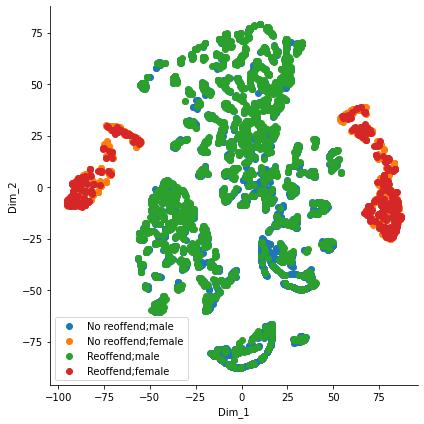}
 \caption{Baseline Compas data}
 \label{fig:fig3}
 \end{subfigure}
 
 \begin{subfigure}[b]{0.25\textwidth}
 \centering
 \includegraphics[width=\textwidth]{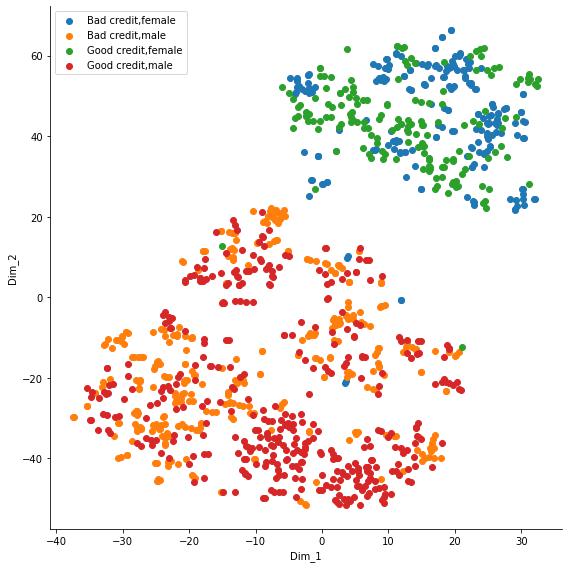}
 \caption{German Credit data oversampled with SMOTE}
 \label{fig:fig1b}
 \end{subfigure}
 \hfill
 \begin{subfigure}[b]{0.25\textwidth}
 \centering
 \includegraphics[width=\textwidth]{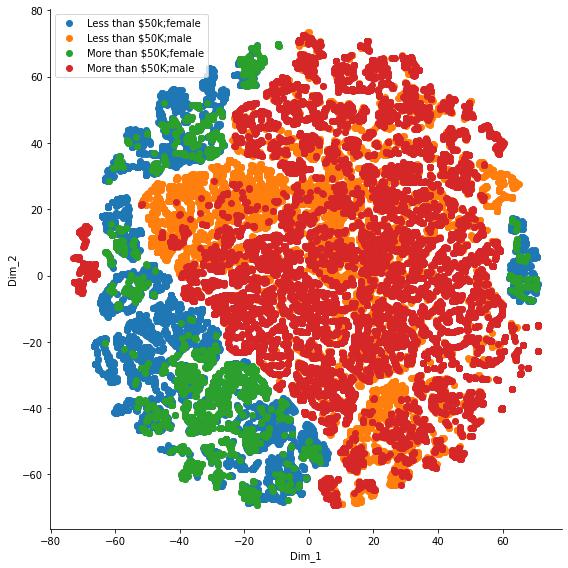}
 \caption{Adult Census data oversampled with SMOTE}
 \label{fig:fig2b}
 \end{subfigure}
 \hfill
 \begin{subfigure}[b]{0.25\textwidth}
 \centering
 \includegraphics[width=\textwidth]{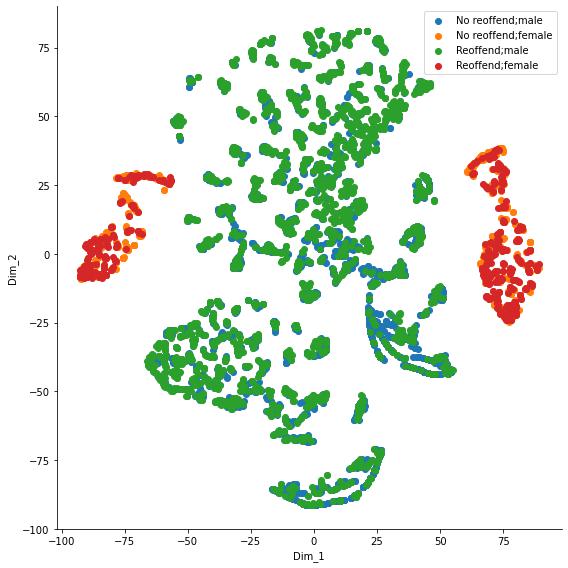}
 \caption{Compas data oversampled with SMOTE}
 \label{fig:fig3b}
 \end{subfigure}

 \begin{subfigure}[b]{0.25\textwidth}
 \centering
 \includegraphics[width=\textwidth]{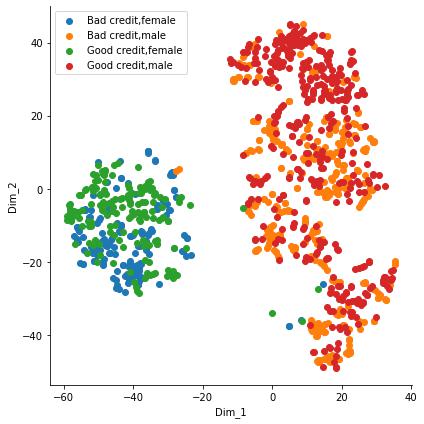}
 \caption{German Credit data oversampled with FOS}
 \label{fig:fig1a}
 \end{subfigure}
 \hfill
 \begin{subfigure}[b]{0.25\textwidth}
 \centering
 \includegraphics[width=\textwidth]{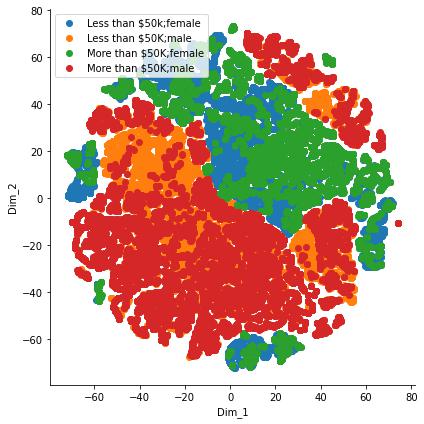}
 \caption{Adult Census data oversampled with FOS}
 \label{fig:fig2a}
 \end{subfigure}
 \hfill
 \begin{subfigure}[b]{0.25\textwidth}
 \centering
 \includegraphics[width=\textwidth]{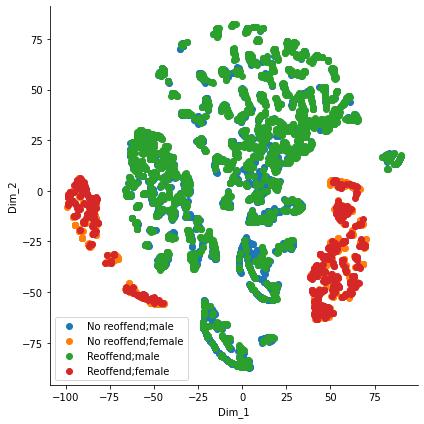}
 \caption{Compas data oversampled with FOS}
 \label{fig:fig3a}
 \end{subfigure}
 
 \caption{The training datasets used in our experiments are split into four groups based on class and protected attributes and visualized with t-SNE. The under-represented groups are oversampled by FOS, resulting in more instances for the respective sensitive feature in the minority class. FOS reduces under-representation by protected groups and aids in blurring distinctions between sensitive features, thus reducing bias. See section 5.4 for discussion.}
 \label{fig:fig4}
\end{figure*}

\begin{figure}
 \centering
 \begin{subfigure}[b]{0.3\textwidth}
 \centering
 \includegraphics[width=\textwidth]{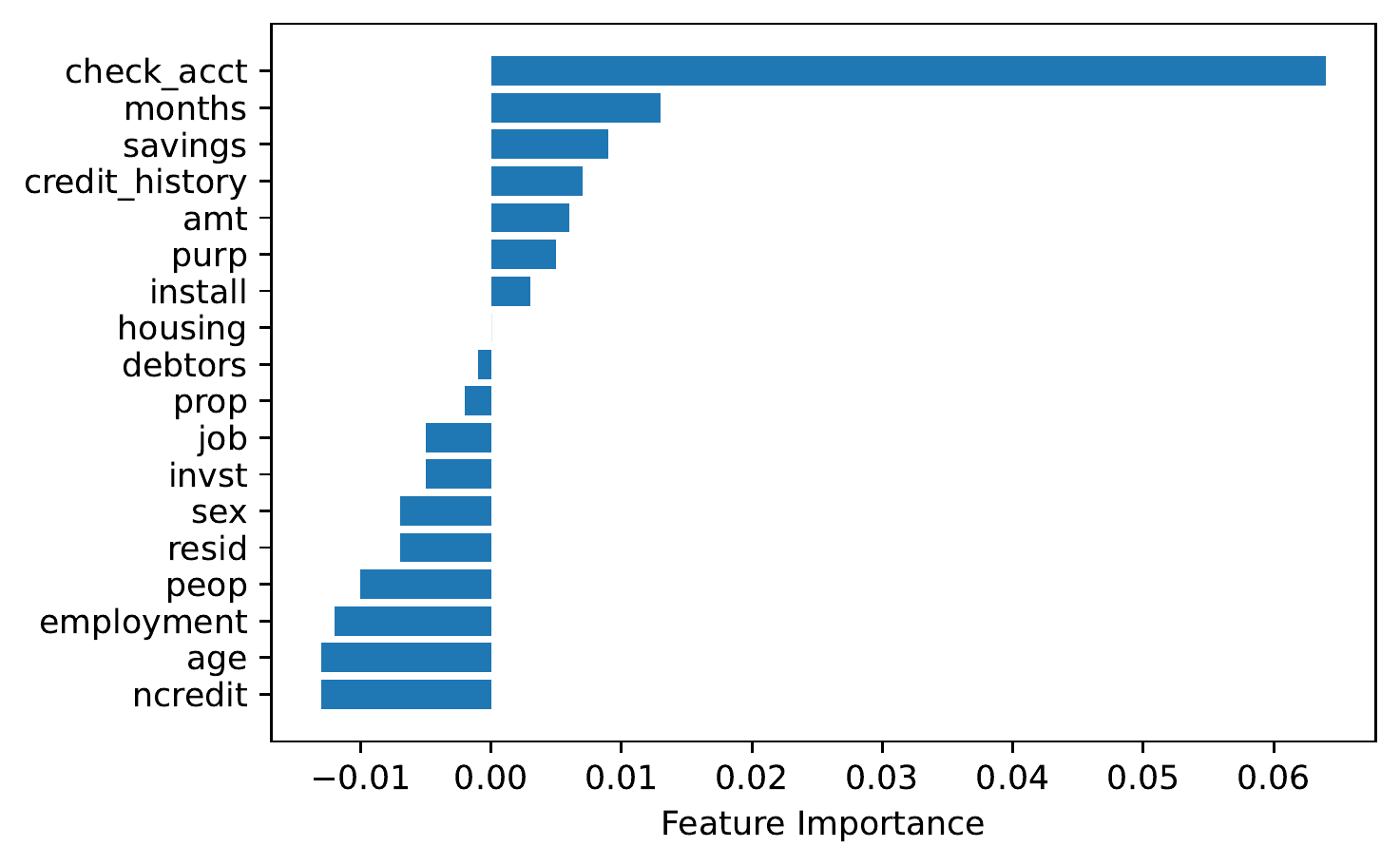}
 \caption{German features}
 \label{fig:fig18}
 \end{subfigure}
 \hfill
 \begin{subfigure}[b]{0.3\textwidth}
 \centering
 \includegraphics[width=\textwidth]{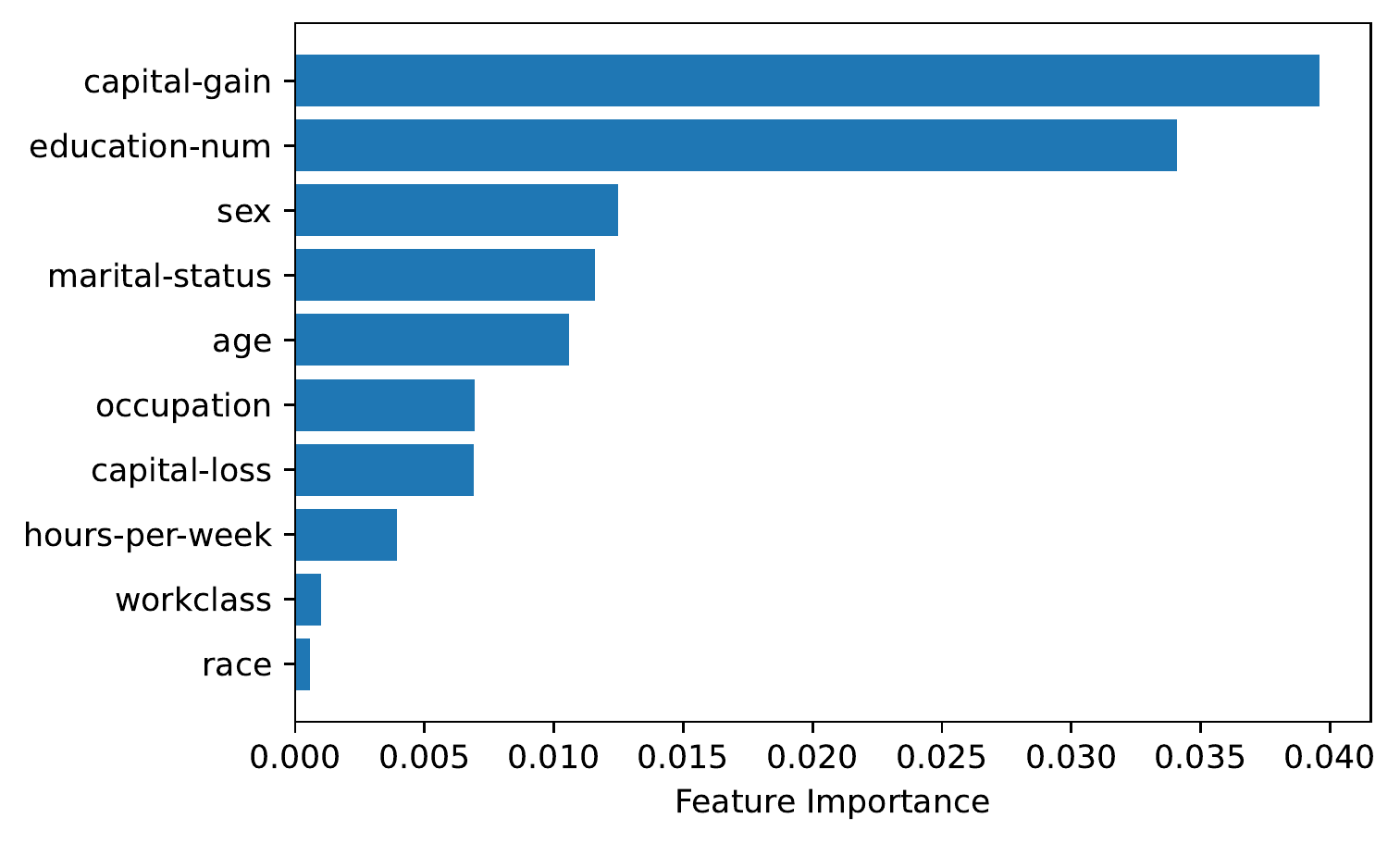}
 \caption{Adult SVM features}
 \label{fig:fig19}
 \end{subfigure}
 \hfill
 \begin{subfigure}[b]{0.3\textwidth}
 \centering
 \includegraphics[width=\textwidth]{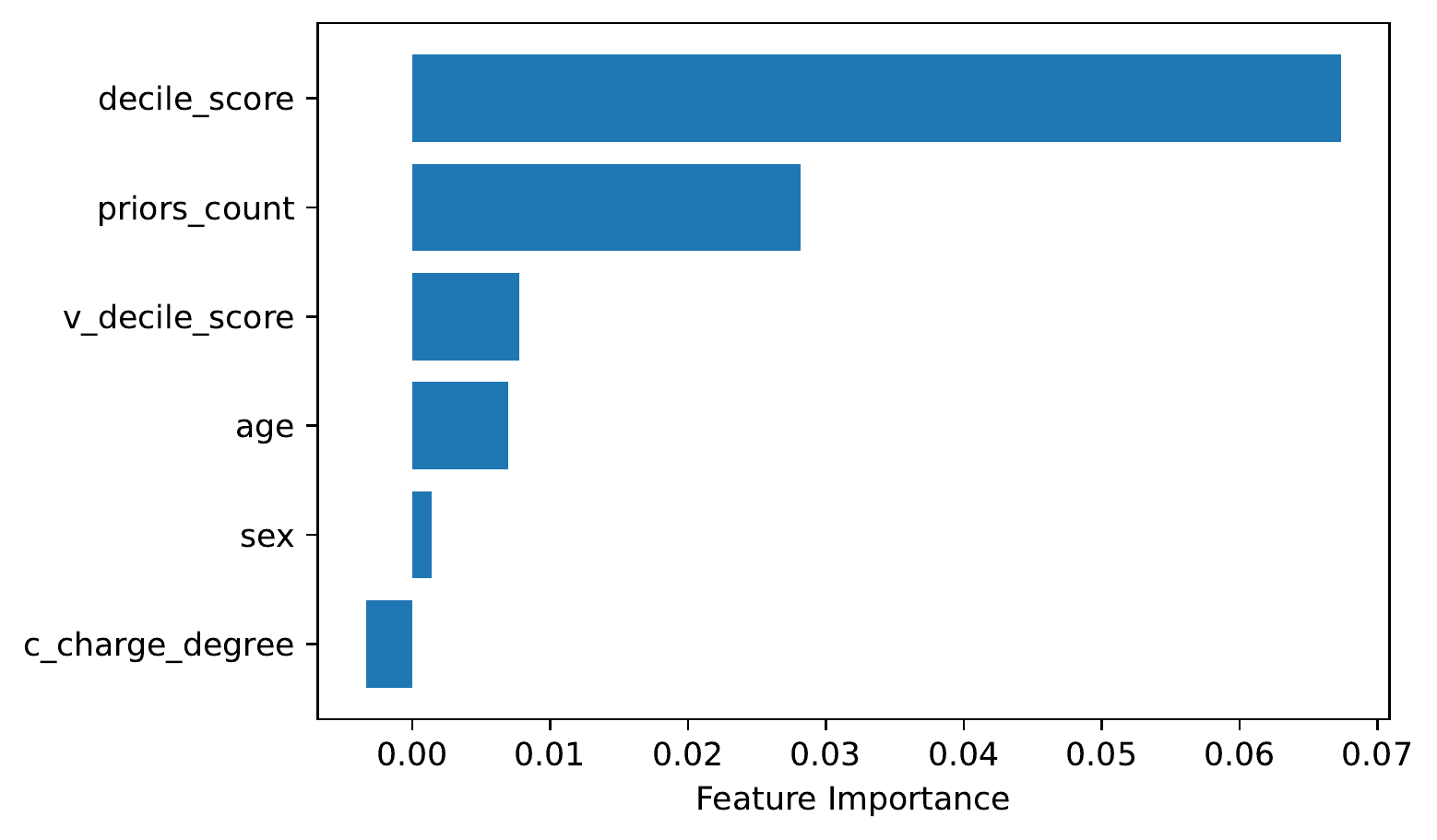}
 \caption{Compas features}
 \label{fig:fig20}
 \end{subfigure}
 
 \begin{subfigure}[b]{0.3\textwidth}
 \centering
 \includegraphics[width=\textwidth]{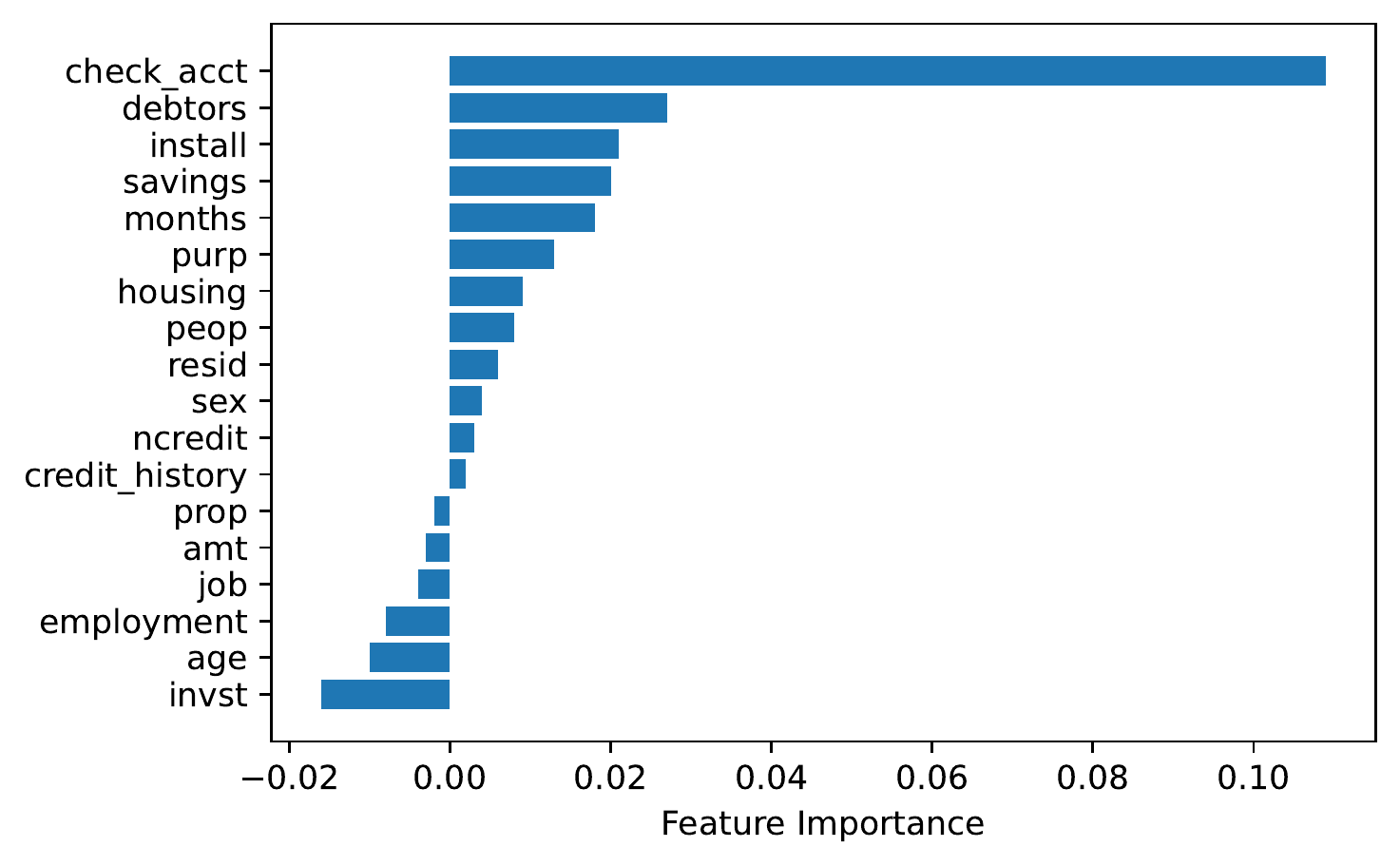}
 \caption{German FOS features}
 \label{fig:fig21}
 \end{subfigure}
 \hfill
 \begin{subfigure}[b]{0.3\textwidth}
 \centering
 \includegraphics[width=\textwidth]{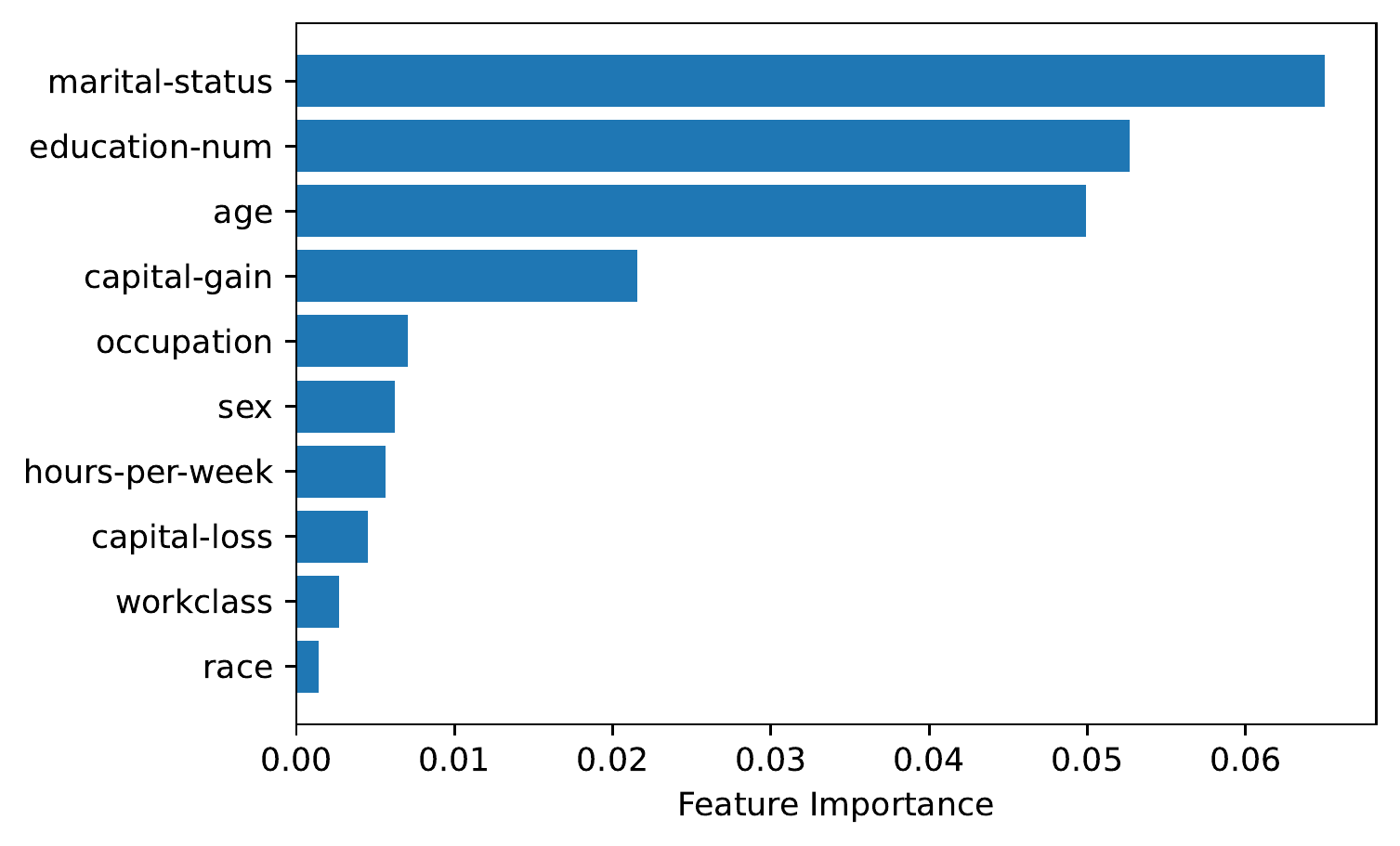}
 \caption{Adult FOS features}
 \label{fig:fig22}
 \end{subfigure}
 \hfill
 \begin{subfigure}[b]{0.3\textwidth}
 \centering
 \includegraphics[width=\textwidth]{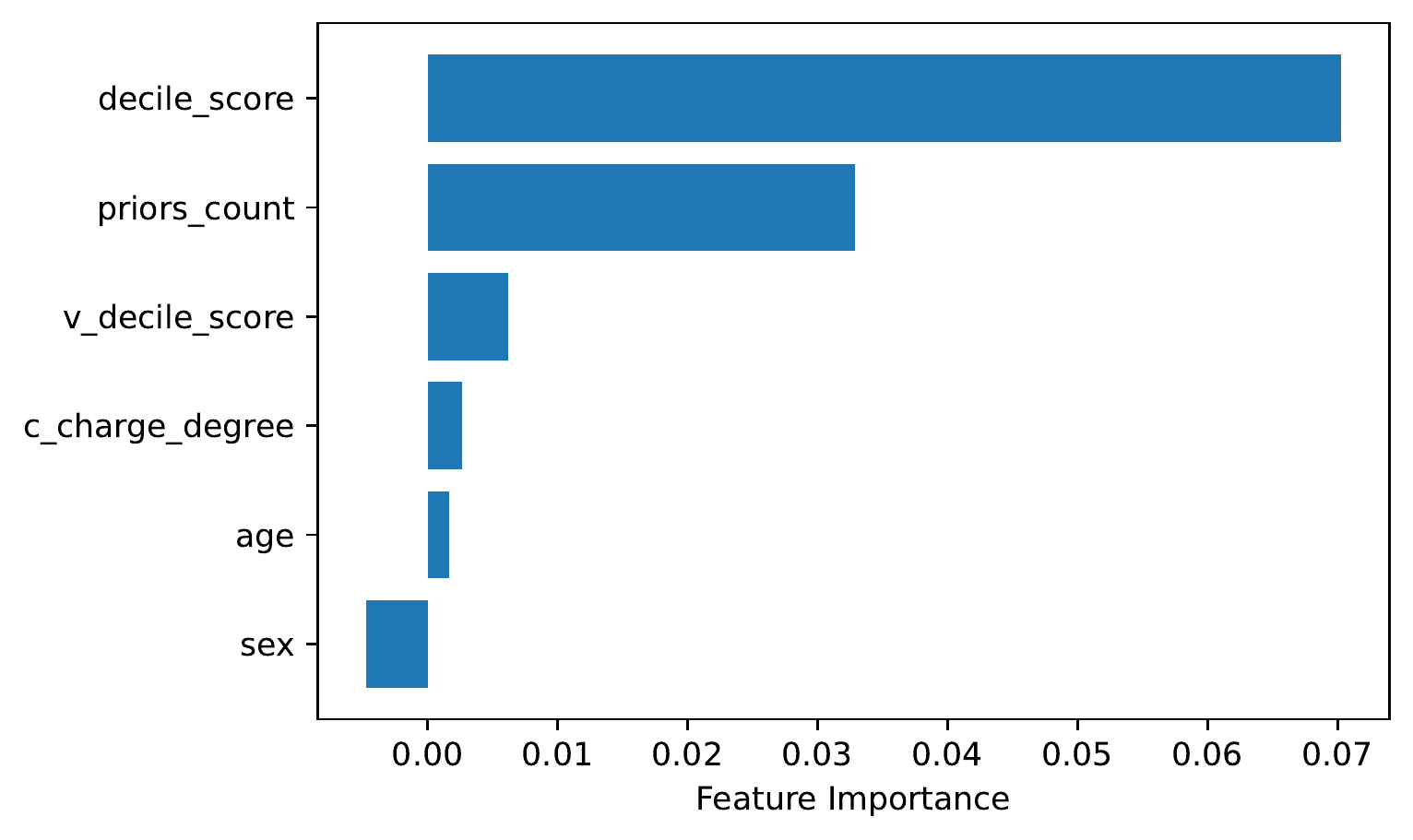}
 \caption{Compas FOS features}
 \label{fig:fig23}
 \end{subfigure}

 \caption{Comparison of SVM feature importance with and without FOS. In the case of the German and Adult datasets, FOS induces a change in feature selection.}
 \label{fig:Features}
\end{figure}

\subsection{Experiment 4: Impact of oversampling on feature importance} 

For the German Credit and Adult datasets, FOS induces some changes in the classifier's feature selection and feature importance. For Compas, FOS does not trigger much of a change in feature selection, likely because it has fewer features than the other two datasets. See Figure~\ref{fig:Features}. For German Credit, FOS increases the importance of the checking account feature (from .06 to .10) and moves the debtors and installment credit features into second and third place. Feature weighting affects a classifier's decision boundary, which in turn, can impact fairness and accuracy. We can see in Figure~\ref{fig:fig4} that FOS does change the placement of training examples. Figure~\ref{fig:fig4} is a two dimensional depiction of the feature space based on t-SNE \citep{van2008visualizing}. FOS generates additional synthetic examples of minority class, females (blue) and minority class, males (orange), as compared to the baseline image. We note that a sub-group of minority and majority females (blue and green) like closer to males (orange and red) and that there is also a small sub-group of males that lie closer to the female group in the FOS t-SNE image. Interestingly, both the baseline and FOS data show a more clear delineation between protected groups (females - blue and green vs. males - orange and red) than classes, which may imply underlying bias in the training data. For Adult Census, FOS changes the importance of the marital status, capital gain, age, sex and years of education features. See Figure~\ref{fig:Features}. Based on its t-SNE, the arrangement of the Adult Census data appears more complex and less linearly separable. It should be noted that, based on its t-SNE, FOS appears to create more sub-groups of females (blue and green) that are embedded in the male cluster (6 clusters with FOS vs. only 2 in the baseline data). This additional sub-group creation may blur the separability between males and females, which causes a classifier to be less certain of the distinction between members of a protected group, thus reducing discrimination (\textbf{RQ4 answered}).

\section{Discussion and lessons learned} 

\noindent \textbf{Discussion.} Several papers have discussed the relationship between fairness and class imbalance in machine learning. Subramanian et al. note that there has traditionally been a disconnect between imbalanced learning and bias reduction; although they believe that the two areas share many similarities \citep{subramanian2021fairness}. They propose a margin-loss approach for tweet sentiment and occupation classification. Ferrari and Bacciu propose a cost-sensitive approach to deal with both imbalanced data and fairness \citep{ferrari2021addressing}. Iosifidis and Ntoutsi discuss imbalanced learning and fairness in the data streaming context \citep{iosifidis2020mathsf} and also consider using data augmentation techniques to increase the number of training instances for under-represented protected groups \citep{iosifidis2018dealing}.

\smallskip
\noindent \textbf{Lessons learned.} Our survey of existing approaches for algorithmic fairness and imbalanced data, as well as experiments conducted to evaluate the proposed FOS approach lead us to learn the following lessons:

\begin{itemize}
    \item \textbf{Two sides of the same coin.} Fairness and class imbalance focus on targeting one specific type of bias present either in data or algorithms, addressing similar issues from different perspectives. However, these two domains are not disjoint and in fact should be considered as mutually complementary. Training data can be imbalanced with respect to classes \textit{and} protected features. For example, the German Credit, Adult Census and Compas datasets (see Table~\ref{tab: dimbal}) all exhibit imbalance with respect to classes and sensitive groups. Discrimination can result from the under-representation of protected groups in training data. 
    
    \item \textbf{Data imbalance techniques can help make standard algorithms more fair.} Our experimental study shows that using a technique derived from the imbalanced domain (oversampling) with modifications ensuring its focus on fairness (see FOS description) can make standard classifiers fairness-aware much more efficiently than using several existing preprocessing techniques taken from either imbalanced or fairness domains. This shows the massive potential in bridging those two domains and creating approaches that leverage and merge state-of-the-art advancements in each of them. 
    
    \item \textbf{Data imbalance techniques can further boost fairness-aware algorithms.} Our experiments additionally proved that by using techniques derived from imbalanced domain, we can significantly improve existing fairness-aware classifiers. This shows that multiple biases may exist in data at once and only by addressing the ways in which both data and algorithms may be skewed can we obtain truly fair machine learning. 
    
    \item \textbf{Robustness and fairness should be seen as a holistic task.} Based on our previous observation, we postulate that robustness and fairness should be seen as a holistic task. When evaluating algorithms, we need to use practices from both domains. Our experiments show the importance of using fairness metrics combined with data-level difficulties, such as an increasing imbalance ratio. Additionally, understanding the impact of features on the classifier’s decisions is crucial to evaluating if protected features truly have no impact on a classifier. This can also be seen as a first step towards explainable systems that shed light on both fairness and class imbalance.  

\end{itemize}

\section{Open challenges and future directions}

Next, we formulate the open challenges and future directions that may be taken by researchers to further develop algorithms that are fair and robust by merging algorithmic fairness with class imbalance approaches. 

\begin{itemize}
    \item \textbf{Unified nomenclature for fairness and imbalance.} Based on our review of the fairness and imbalanced learning literature, we can see that although some of the nomenclature varies between these fields, the basic methodology is similar. For example, both imbalanced learning and fairness utilize three basic approaches: (1) pre-processing / data-level methods, (2) in-processing / algorithm level methods, and (3) post-processing / hybrid methods. Common metrics used in imbalanced learning and group fairness also are derived from a binary confusion matrix \citep{johnson2019survey}. Therefore, there is a need for unified nomenclature accepted by researchers from both fields that will enable and foster collaboration between these two domains.
    
    \item \textbf{Holistic and reproducible experimental framework.} The algorithmic fairness domain could significantly benefit from incorporating similar rigorous experimental protocols that are nowadays becoming standard in imbalanced learning \citep{Fernandez:2018}. There is a need for agreed upon usage of benchmarks, performance metrics, baseline models, as well as verifying obtained results with statistical analysis \citep{Stapor:2021}. By creating a widely accepted experimental testbed, the new methods that aim at countering various biases in data can be truly fairly analyzed and compared with state-of-the-art. Furthermore, a software library offering easy to use and an extensible set of the most effective methods for fairness and imbalance-aware methods should be developed. 
    
    \item \textbf{Exploration of different levels of interplay between fairness and imbalance domains.} In this work, we showed how oversampling can be used not only to achieve fairness in classification, but also further boost fairness-aware classifiers. As the imbalanced learning literature is rich with diverse ideas on how to counter bias in algorithms\citep{Fernandez:2018}, we anticipate a great potential in applying other families of imbalanced approaches to algorithmic fairness. Cost-sensitive \citep{ZhangT:2019} and skew-insensitive \citep{Wen:2021} mechanisms stand out as the most obvious choices, but we anticipate that other methods may prove beneficial, such as analyzing the usefulness of instance-level properties of minority classes and their impact on fairness \citep{Skryjomski:2017}, learning discriminative and fair low-dimensional representations \citep{Korycki:2021ld}, or using one-class classification tailored for protected features \citep{Bellinger:2017}. 
    
    \item \textbf{Robustness to adversarial attacks.} The lack of fairness in machine learning may not only be caused by data or algorithms themselves. Decision-making systems may be subject to adversarial attacks caused by a malicious party that aims at damaging fairness to either discredit a given system or counter the progress towards more transparent and just society \citep{Yeom:2020}. We see developing robust mechanisms for fairness-aware algorithms as a crucial step towards fighting bias \citep{Xu:2021}. This calls for understanding how fairness-aware algorithms can be poisoned by attacks, creating unified benchmarks with adversarial attacks targeting fairness, and developing adversarial training protocols to prepare robust learning systems to be deployed in the field. 
    
    \item \textbf{Fairness and robustness to imbalance in continual and streaming environments.} All of the works discussed in this paper considered static classification problems. However, modern data sources constantly generate new information, leading to the need for learning from data streams and continual learning. The former focuses on adaptation to changes, known as concept drift \citep{Korycki:2021cvpr}, while the latter focuses on retaining previously learned information and avoiding catastrophic forgetting \citep{Parisi:2019}. Considering algorithmic fairness in such dynamic scenarios can lead to a plethora of exciting challenges, such as the evolving nature of bias in data, the emergence of new protected attributes over time, or selective forgetting and adaptation towards boosting fairness and reducing bias over time \citep{Aguiar:2022}. 
    
    \item \textbf{XAI for fairness and robustness to imbalance.} Explainable artificial intelligence allows end-users to understand the decision-making process of a classifier, as well as gain insights and better understanding of the analyzed data properties \citep{Lin:2021}. This is crucial from the perspective of algorithmic fairness, as XAI should lead to transparency with respect to how a classifier was able to avoid bias \citep{Greeff:2021}. Furthermore, XAI may allow for discovering new, unknown biases in both data and the learning model itself \citep{Fu:2020}. XAI has not been sufficiently explored in the context of imbalanced data, thus making it an important open challenge. Creating tools for the explainable understanding of various types of bias will allow us to further bridge these two domains by better comprehending the interplay between them. 
    
\end{itemize}

\section{Conclusion}
A key facet in reducing algorithmic discrimination with respect to protected groups is the simultaneous reduction of class and protected group imbalance in training data. We show that reducing data imbalance facilitates improvements in both model accuracy and group fairness. This paper discussed the importance of bridging imbalanced learning and group fairness, by showing how key concepts in these fields overlap, and it proposed a novel oversampling algorithm, \emph{Fair Oversampling}, that addresses both skewed class distributions and protected attributes. Our method: (i) can be used as an efficient pre-processing algorithm with standard ML algorithms to jointly improve imbalance and fairness; and (ii) can supplement fairness-aware learning algorithms to improve their robustness. Additionally, we take a step toward bridging the gap between fairness and imbalanced learning with a new metric, \emph{Fair Utility}, that combines balanced accuracy with group fairness measures.


\bibliographystyle{spbasic} 
\bibliography{paper_bib}

\begin{thebibliography}{88}
\providecommand{\natexlab}[1]{#1}
\providecommand{\url}[1]{{#1}}
\providecommand{\urlprefix}{URL }
\expandafter\ifx\csname urlstyle\endcsname\relax
  \providecommand{\doi}[1]{DOI~\discretionary{}{}{}#1}\else
  \providecommand{\doi}{DOI~\discretionary{}{}{}\begingroup
  \urlstyle{rm}\Url}\fi
\providecommand{\eprint}[2][]{\url{#2}}

\bibitem[{Agarwal et~al(2018)Agarwal, Beygelzimer, Dud{\'\i}k, Langford, and
  Wallach}]{agarwal2018reductions}
Agarwal A, Beygelzimer A, Dud{\'\i}k M, Langford J, Wallach H (2018) A
  reductions approach to fair classification. In: International Conference on
  Machine Learning, PMLR, pp 60--69

\bibitem[{Aguiar et~al(2022)Aguiar, Krawczyk, and Cano}]{Aguiar:2022}
Aguiar G, Krawczyk B, Cano A (2022) A survey on learning from imbalanced data
  streams: taxonomy, challenges, empirical study, and reproducible experimental
  framework. CoRR abs/2204.03719, \doi{10.48550/arXiv.2204.03719},
  \urlprefix\url{https://doi.org/10.48550/arXiv.2204.03719},
  \eprint{2204.03719}

\bibitem[{Angwin(2016)}]{propublic}
Angwin J (2016) Machine bias: There's software used across the country to
  predict future criminals and its biased against blacks.
  \urlprefix\url{https://github.com/Trusted-AI/AIF360/blob/master/aif360/data/raw/compas/README.md}

\bibitem[{Barocas and Selbst(2016)}]{barocas2016big}
Barocas S, Selbst AD (2016) Big data's disparate impact. Calif L Rev 104:671

\bibitem[{Bej et~al(2021)Bej, Davtyan, Wolfien, Nassar, and
  Wolkenhauer}]{Bej:2021}
Bej S, Davtyan N, Wolfien M, Nassar M, Wolkenhauer O (2021) Loras: an
  oversampling approach for imbalanced datasets. Mach Learn 110(2):279--301

\bibitem[{Bellamy et~al(2018)Bellamy, Dey, Hind, Hoffman, Houde, Kannan, Lohia,
  Martino, Mehta, Mojsilovic et~al}]{bellamy2018ai}
Bellamy RK, Dey K, Hind M, Hoffman SC, Houde S, Kannan K, Lohia P, Martino J,
  Mehta S, Mojsilovic A, et~al (2018) Ai fairness 360: An extensible toolkit
  for detecting, understanding, and mitigating unwanted algorithmic bias. arXiv
  preprint arXiv:181001943

\bibitem[{Bellinger et~al(2017)Bellinger, Sharma, Za{\"{\i}}ane, and
  Japkowicz}]{Bellinger:2017}
Bellinger C, Sharma S, Za{\"{\i}}ane OR, Japkowicz N (2017) Sampling a longer
  life: Binary versus one-class classification revisited. In: First
  International Workshop on Learning with Imbalanced Domains: Theory and
  Applications, LIDTA@PKDD/ECML 2017, 22 September 2017, Skopje, Macedonia,
  {PMLR}, Proceedings of Machine Learning Research, vol~74, pp 64--78

\bibitem[{Berk et~al(2021)Berk, Heidari, Jabbari, Kearns, and
  Roth}]{berk2021fairness}
Berk R, Heidari H, Jabbari S, Kearns M, Roth A (2021) Fairness in criminal
  justice risk assessments: The state of the art. Sociological Methods \&
  Research 50(1):3--44

\bibitem[{Biswas and Rajan(2020)}]{biswas2020machine}
Biswas S, Rajan H (2020) Do the machine learning models on a crowd sourced
  platform exhibit bias? an empirical study on model fairness. In: Proceedings
  of the 28th ACM Joint Meeting on European Software Engineering Conference and
  Symposium on the Foundations of Software Engineering, pp 642--653

\bibitem[{Bonchi et~al(2017)Bonchi, Hajian, Mishra, and
  Ramazzotti}]{bonchi2017exposing}
Bonchi F, Hajian S, Mishra B, Ramazzotti D (2017) Exposing the probabilistic
  causal structure of discrimination. International Journal of Data Science and
  Analytics 3(1):1--21

\bibitem[{Bunkhumpornpat et~al(2009)Bunkhumpornpat, Sinapiromsaran, and
  Lursinsap}]{bunkhumpornpat2009safe}
Bunkhumpornpat C, Sinapiromsaran K, Lursinsap C (2009) Safe-level-smote:
  Safe-level-synthetic minority over-sampling technique for handling the class
  imbalanced problem. In: Pacific-Asia conference on knowledge discovery and
  data mining, Springer, pp 475--482

\bibitem[{Calders and Verwer(2010)}]{calders2010three}
Calders T, Verwer S (2010) Three naive bayes approaches for discrimination-free
  classification. Data mining and knowledge discovery 21(2):277--292

\bibitem[{Caliskan et~al(2017)Caliskan, Bryson, and
  Narayanan}]{caliskan2017semantics}
Caliskan A, Bryson JJ, Narayanan A (2017) Semantics derived automatically from
  language corpora contain human-like biases. Science 356(6334):183--186

\bibitem[{Calmon et~al(2017)Calmon, Wei, Vinzamuri, Ramamurthy, and
  Varshney}]{calmon2017optimized}
Calmon FP, Wei D, Vinzamuri B, Ramamurthy KN, Varshney KR (2017) Optimized
  pre-processing for discrimination prevention. In: Proceedings of the 31st
  International Conference on Neural Information Processing Systems, pp
  3995--4004

\bibitem[{Cao et~al(2019)Cao, Wei, Gaidon, Arechiga, and Ma}]{cao2019learning}
Cao K, Wei C, Gaidon A, Arechiga N, Ma T (2019) Learning imbalanced datasets
  with label-distribution-aware margin loss. arXiv preprint arXiv:190607413

\bibitem[{Chakraborty et~al(2021)Chakraborty, Majumder, and
  Menzies}]{chakraborty2021bias}
Chakraborty J, Majumder S, Menzies T (2021) Bias in machine learning software:
  Why? how? what to do? arXiv preprint arXiv:210512195

\bibitem[{Chawla et~al(2002)Chawla, Bowyer, Hall, and
  Kegelmeyer}]{chawla2002smote}
Chawla NV, Bowyer KW, Hall LO, Kegelmeyer WP (2002) Smote: synthetic minority
  over-sampling technique. Journal of artificial intelligence research
  16:321--357

\bibitem[{Chouldechova(2017)}]{chouldechova2017fair}
Chouldechova A (2017) Fair prediction with disparate impact: A study of bias in
  recidivism prediction instruments. Big data 5(2):153--163

\bibitem[{Chouldechova et~al(2018)Chouldechova, Benavides-Prado, Fialko, and
  Vaithianathan}]{chouldechova2018case}
Chouldechova A, Benavides-Prado D, Fialko O, Vaithianathan R (2018) A case
  study of algorithm-assisted decision making in child maltreatment hotline
  screening decisions. In: Conference on Fairness, Accountability and
  Transparency, PMLR, pp 134--148

\bibitem[{Cieslak et~al(2006)Cieslak, Chawla, and
  Striegel}]{cieslak2006combating}
Cieslak DA, Chawla NV, Striegel A (2006) Combating imbalance in network
  intrusion datasets. In: GrC, Citeseer, pp 732--737

\bibitem[{Commission(2021)}]{eu}
Commission E (2021) Proposal for a regulation laying down harmonised rules on
  artificial intelligence (artificial intelligence act).
  \urlprefix\url{https://eur-lex.europa.eu/legal-content/EN/TXT/?uri=CELEX\%3A52021PC0206}

\bibitem[{Corbett-Davies and Goel(2018)}]{corbett2018measure}
Corbett-Davies S, Goel S (2018) The measure and mismeasure of fairness: A
  critical review of fair machine learning. arXiv preprint arXiv:180800023

\bibitem[{Corbett-Davies et~al(2017)Corbett-Davies, Pierson, Feller, Goel, and
  Huq}]{corbett2017algorithmic}
Corbett-Davies S, Pierson E, Feller A, Goel S, Huq A (2017) Algorithmic
  decision making and the cost of fairness. In: Proceedings of the 23rd acm
  sigkdd international conference on knowledge discovery and data mining, pp
  797--806

\bibitem[{Cui et~al(2019)Cui, Jia, Lin, Song, and Belongie}]{cui2019class}
Cui Y, Jia M, Lin TY, Song Y, Belongie S (2019) Class-balanced loss based on
  effective number of samples. In: Proceedings of the IEEE/CVF conference on
  computer vision and pattern recognition, pp 9268--9277

\bibitem[{Dablain et~al(2022)Dablain, Krawczyk, and
  Chawla}]{dablain2021deepsmote}
Dablain D, Krawczyk B, Chawla NV (2022) Deep{SMOTE}: Fusing deep learning and
  smote for imbalanced data. IEEE Transactions on Neural Networks and Learning
  Systems pp 1--15, \doi{10.1109/TNNLS.2021.3136503}

\bibitem[{Dastin(2018)}]{reuters}
Dastin J (2018) Amazon scraps secret ai recruiting tool that showed bias
  against women.
  \urlprefix\url{https://www.reuters.com/article/us-amazon-com-jobs-automation-insight/amazon-scraps-secret-ai-recruiting-tool-that-showed-bias-against-women-idUSKCN1MK08G}

\bibitem[{Dwork et~al(2012)Dwork, Hardt, Pitassi, Reingold, and
  Zemel}]{dwork2012fairness}
Dwork C, Hardt M, Pitassi T, Reingold O, Zemel R (2012) Fairness through
  awareness. In: Proceedings of the 3rd innovations in theoretical computer
  science conference, pp 214--226

\bibitem[{Edwards and Storkey(2015)}]{edwards2015censoring}
Edwards H, Storkey A (2015) Censoring representations with an adversary. arXiv
  preprint arXiv:151105897

\bibitem[{Feldman et~al(2015)Feldman, Friedler, Moeller, Scheidegger, and
  Venkatasubramanian}]{feldman2015certifying}
Feldman M, Friedler SA, Moeller J, Scheidegger C, Venkatasubramanian S (2015)
  Certifying and removing disparate impact. In: proceedings of the 21th ACM
  SIGKDD international conference on knowledge discovery and data mining, pp
  259--268

\bibitem[{Fern{\'{a}}ndez et~al(2018)Fern{\'{a}}ndez, Garc{\'{\i}}a, Galar,
  Prati, Krawczyk, and Herrera}]{Fernandez:2018}
Fern{\'{a}}ndez A, Garc{\'{\i}}a S, Galar M, Prati RC, Krawczyk B, Herrera F
  (2018) Learning from Imbalanced Data Sets. Springer,
  \doi{10.1007/978-3-319-98074-4},
  \urlprefix\url{https://doi.org/10.1007/978-3-319-98074-4}

\bibitem[{Ferrari and Bacciu(2021)}]{ferrari2021addressing}
Ferrari E, Bacciu D (2021) Addressing fairness, bias and class imbalance in
  machine learning: the fbi-loss. arXiv preprint arXiv:210506345

\bibitem[{Flores et~al(2016)Flores, Bechtel, and Lowenkamp}]{flores2016false}
Flores AW, Bechtel K, Lowenkamp CT (2016) False positives, false negatives, and
  false analyses: A rejoinder to machine bias: There's software used across the
  country to predict future criminals. and it's biased against blacks. Fed
  Probation 80:38

\bibitem[{Fu et~al(2020)Fu, Xian, Gao, Zhao, Huang, Ge, Xu, Geng, Shah, Zhang,
  and de~Melo}]{Fu:2020}
Fu Z, Xian Y, Gao R, Zhao J, Huang Q, Ge Y, Xu S, Geng S, Shah C, Zhang Y,
  de~Melo G (2020) Fairness-aware explainable recommendation over knowledge
  graphs. In: Proceedings of the 43rd International {ACM} {SIGIR} conference on
  research and development in Information Retrieval, {SIGIR} 2020, Virtual
  Event, China, July 25-30, 2020, {ACM}, pp 69--78

\bibitem[{de~Greeff et~al(2021)de~Greeff, de~Boer, Hillerstr{\"{o}}m, Bomhof,
  Jorritsma, and Neerincx}]{Greeff:2021}
de~Greeff J, de~Boer MHT, Hillerstr{\"{o}}m FHJ, Bomhof F, Jorritsma W,
  Neerincx MA (2021) The {FATE} system: Fair, transparent and explainable
  decision making. In: Proceedings of the {AAAI} 2021 Spring Symposium on
  Combining Machine Learning and Knowledge Engineering {(AAAI-MAKE} 2021),
  Stanford University, Palo Alto, California, USA, March 22-24, 2021,
  CEUR-WS.org, {CEUR} Workshop Proceedings, vol 2846

\bibitem[{Grgic-Hlaca et~al(2016)Grgic-Hlaca, Zafar, Gummadi, and
  Weller}]{grgic2016case}
Grgic-Hlaca N, Zafar MB, Gummadi KP, Weller A (2016) The case for process
  fairness in learning: Feature selection for fair decision making. In: NIPS
  symposium on machine learning and the law, vol~1, p~2

\bibitem[{Halevy et~al(2009)Halevy, Norvig, and
  Pereira}]{halevy2009unreasonable}
Halevy A, Norvig P, Pereira F (2009) The unreasonable effectiveness of data.
  IEEE Intelligent Systems 24(2):8--12

\bibitem[{Han et~al(2005)Han, Wang, and Mao}]{han2005borderline}
Han H, Wang WY, Mao BH (2005) Borderline-smote: a new over-sampling method in
  imbalanced data sets learning. In: International conference on intelligent
  computing, Springer, pp 878--887

\bibitem[{Hardt et~al(2016)Hardt, Price, and Srebro}]{hardt2016equality}
Hardt M, Price E, Srebro N (2016) Equality of opportunity in supervised
  learning. Advances in neural information processing systems 29:3315--3323

\bibitem[{Hashimoto et~al(2018)Hashimoto, Srivastava, Namkoong, and
  Liang}]{hashimoto2018fairness}
Hashimoto T, Srivastava M, Namkoong H, Liang P (2018) Fairness without
  demographics in repeated loss minimization. In: International Conference on
  Machine Learning, PMLR, pp 1929--1938

\bibitem[{Hofmann(1994)}]{uciger}
Hofmann DH (1994) Statlog (german credit data) data set.
  \urlprefix\url{https://archive.ics.uci.edu/ml/datasets/Statlog+\%28German+Credit+Data\%29}

\bibitem[{Iosifidis and Ntoutsi(2018)}]{iosifidis2018dealing}
Iosifidis V, Ntoutsi E (2018) Dealing with bias via data augmentation in
  supervised learning scenarios. Jo Bates Paul D Clough Robert J{\"a}schke 24

\bibitem[{Iosifidis and Ntoutsi(2020)}]{iosifidis2020mathsf}
Iosifidis V, Ntoutsi E (2020) Fabboo: Online fairness-aware learning under
  class imbalance. In: International Conference on Discovery Science, Springer,
  pp 159--174

\bibitem[{Johnson and Khoshgoftaar(2019)}]{johnson2019survey}
Johnson JM, Khoshgoftaar TM (2019) Survey on deep learning with class
  imbalance. Journal of Big Data 6(1):1--54

\bibitem[{Kahn and Marshall(1953)}]{kahn1953methods}
Kahn H, Marshall AW (1953) Methods of reducing sample size in monte carlo
  computations. Journal of the Operations Research Society of America
  1(5):263--278

\bibitem[{Kamiran et~al(2013)Kamiran, {\v{Z}}liobait{e}, and
  Calders}]{kamiran2013quantifying}
Kamiran F, {\v{Z}}liobait{e} I, Calders T (2013) Quantifying explainable
  discrimination and removing illegal discrimination in automated decision
  making. Knowledge and information systems 35(3):613--644

\bibitem[{Khandani et~al(2010)Khandani, Kim, and Lo}]{khandani2010consumer}
Khandani AE, Kim AJ, Lo AW (2010) Consumer credit-risk models via
  machine-learning algorithms. Journal of Banking \& Finance 34(11):2767--2787

\bibitem[{Kleinberg(2018)}]{kleinberg2018inherent}
Kleinberg J (2018) Inherent trade-offs in algorithmic fairness. In: Abstracts
  of the 2018 ACM International Conference on Measurement and Modeling of
  Computer Systems, pp 40--40

\bibitem[{Kohavi et~al(1996)}]{kohavi1996scaling}
Kohavi R, et~al (1996) Scaling up the accuracy of naive-bayes classifiers: A
  decision-tree hybrid. In: Kdd, vol~96, pp 202--207

\bibitem[{Korycki and Krawczyk(2021{\natexlab{a}})}]{Korycki:2021cvpr}
Korycki L, Krawczyk B (2021{\natexlab{a}}) Class-incremental experience replay
  for continual learning under concept drift. In: {IEEE} Conference on Computer
  Vision and Pattern Recognition Workshops, {CVPR} Workshops 2021, virtual,
  June 19-25, 2021, Computer Vision Foundation / {IEEE}, pp 3649--3658

\bibitem[{Korycki and Krawczyk(2021{\natexlab{b}})}]{Korycki:2021ld}
Korycki L, Krawczyk B (2021{\natexlab{b}}) Low-dimensional representation
  learning from imbalanced data streams. In: Advances in Knowledge Discovery
  and Data Mining - 25th Pacific-Asia Conference, {PAKDD} 2021, Virtual Event,
  May 11-14, 2021, Proceedings, Part {I}, Springer, Lecture Notes in Computer
  Science, vol 12712, pp 629--641

\bibitem[{Krawczyk(2016)}]{krawczyk2016learning}
Krawczyk B (2016) Learning from imbalanced data: open challenges and future
  directions. Progress in Artificial Intelligence 5(4):221--232

\bibitem[{Krawczyk et~al(2020)Krawczyk, Koziarski, and Wozniak}]{Krawczyk:2020}
Krawczyk B, Koziarski M, Wozniak M (2020) Radial-based oversampling for
  multiclass imbalanced data classification. {IEEE} Trans Neural Networks Learn
  Syst 31(8):2818--2831

\bibitem[{Kubat et~al(1998)Kubat, Holte, and Matwin}]{kubat1998machine}
Kubat M, Holte RC, Matwin S (1998) Machine learning for the detection of oil
  spills in satellite radar images. Machine learning 30(2):195--215

\bibitem[{Lango and Stefanowski(2018)}]{Lango:2018}
Lango M, Stefanowski J (2018) Multi-class and feature selection extensions of
  roughly balanced bagging for imbalanced data. J Intell Inf Syst 50(1):97--127

\bibitem[{Lin et~al(2017)Lin, Goyal, Girshick, He, and
  Doll{\'a}r}]{lin2017focal}
Lin TY, Goyal P, Girshick R, He K, Doll{\'a}r P (2017) Focal loss for dense
  object detection. In: Proceedings of the IEEE international conference on
  computer vision, pp 2980--2988

\bibitem[{Lin et~al(2021)Lin, Lee, and Celik}]{Lin:2021}
Lin Y, Lee W, Celik ZB (2021) What do you see?: Evaluation of explainable
  artificial intelligence {(XAI)} interpretability through neural backdoors.
  In: Zhu F, Ooi BC, Miao C (eds) {KDD} '21: The 27th {ACM} {SIGKDD} Conference
  on Knowledge Discovery and Data Mining, Virtual Event, Singapore, August
  14-18, 2021, {ACM}, pp 1027--1035

\bibitem[{Van~der Maaten and Hinton(2008)}]{van2008visualizing}
Van~der Maaten L, Hinton G (2008) Visualizing data using t-sne. Journal of
  machine learning research 9(11)

\bibitem[{Narayanan(2018)}]{narayanan2018translation}
Narayanan A (2018) Translation tutorial: 21 fairness definitions and their
  politics. In: Proc. Conf. Fairness Accountability Transp., New York, USA, vol
  1170

\bibitem[{Parisi et~al(2019)Parisi, Kemker, Part, Kanan, and
  Wermter}]{Parisi:2019}
Parisi GI, Kemker R, Part JL, Kanan C, Wermter S (2019) Continual lifelong
  learning with neural networks: {A} review. Neural Networks 113:54--71

\bibitem[{Quy et~al(2021)Quy, Roy, Iosifidis, and Ntoutsi}]{Quy:2021}
Quy TL, Roy A, Iosifidis V, Ntoutsi E (2021) A survey on datasets for
  fairness-aware machine learning. CoRR abs/2110.00530,
  \urlprefix\url{https://arxiv.org/abs/2110.00530}, \eprint{2110.00530}

\bibitem[{Raji et~al(2020)Raji, Gebru, Mitchell, Buolamwini, Lee, and
  Denton}]{raji2020saving}
Raji ID, Gebru T, Mitchell M, Buolamwini J, Lee J, Denton E (2020) Saving face:
  Investigating the ethical concerns of facial recognition auditing. In:
  Proceedings of the AAAI/ACM Conference on AI, Ethics, and Society, pp
  145--151

\bibitem[{Rao et~al(2006)Rao, Krishnan, and Niculescu}]{rao2006data}
Rao RB, Krishnan S, Niculescu RS (2006) Data mining for improved cardiac care.
  Acm Sigkdd Explorations Newsletter 8(1):3--10

\bibitem[{Rawls(2001)}]{rawls2001justice}
Rawls J (2001) Justice as fairness: A restatement. Harvard University Press

\bibitem[{Ridnik et~al(2021)Ridnik, Ben-Baruch, Zamir, Noy, Friedman, Protter,
  and Zelnik-Manor}]{ridnik2021asymmetric}
Ridnik T, Ben-Baruch E, Zamir N, Noy A, Friedman I, Protter M, Zelnik-Manor L
  (2021) Asymmetric loss for multi-label classification. In: Proceedings of the
  IEEE/CVF International Conference on Computer Vision, pp 82--91

\bibitem[{Romei and Ruggieri(2014)}]{romei2014multidisciplinary}
Romei A, Ruggieri S (2014) A multidisciplinary survey on discrimination
  analysis. The Knowledge Engineering Review 29(5):582--638

\bibitem[{Schumann et~al(2020)Schumann, Foster, Mattei, and
  Dickerson}]{schumann2020we}
Schumann C, Foster J, Mattei N, Dickerson J (2020) We need fairness and
  explainability in algorithmic hiring. In: International Conference on
  Autonomous Agents and Multi-Agent Systems (AAMAS)

\bibitem[{Seiffert et~al(2007)Seiffert, Khoshgoftaar, Van~Hulse, and
  Napolitano}]{seiffert2007mining}
Seiffert C, Khoshgoftaar TM, Van~Hulse J, Napolitano A (2007) Mining data with
  rare events: a case study. In: 19th IEEE International Conference on Tools
  with Artificial Intelligence (ICTAI 2007), IEEE, vol~2, pp 132--139

\bibitem[{Sharma et~al(2018)Sharma, Bellinger, Krawczyk, Za{\"{\i}}ane, and
  Japkowicz}]{Sharma:2018}
Sharma S, Bellinger C, Krawczyk B, Za{\"{\i}}ane OR, Japkowicz N (2018)
  Synthetic oversampling with the majority class: {A} new perspective on
  handling extreme imbalance. In: {IEEE} International Conference on Data
  Mining, {ICDM} 2018, Singapore, November 17-20, 2018, {IEEE} Computer
  Society, pp 447--456

\bibitem[{Shroff(2017)}]{shroff2017predictive}
Shroff R (2017) Predictive analytics for city agencies: Lessons from children's
  services. Big data 5(3):189--196

\bibitem[{Skryjomski and Krawczyk(2017)}]{Skryjomski:2017}
Skryjomski P, Krawczyk B (2017) Influence of minority class instance types on
  {SMOTE} imbalanced data oversampling. In: First International Workshop on
  Learning with Imbalanced Domains: Theory and Applications, LIDTA@PKDD/ECML
  2017, 22 September 2017, Skopje, Macedonia, {PMLR}, Proceedings of Machine
  Learning Research, vol~74, pp 7--21

\bibitem[{Sleeman and Krawczyk(2021)}]{Sleeman:2021}
Sleeman WC, Krawczyk B (2021) Multi-class imbalanced big data classification on
  spark. Knowl Based Syst 212:106598

\bibitem[{Speicher et~al(2018)Speicher, Heidari, Grgic-Hlaca, Gummadi, Singla,
  Weller, and Zafar}]{speicher2018unified}
Speicher T, Heidari H, Grgic-Hlaca N, Gummadi KP, Singla A, Weller A, Zafar MB
  (2018) A unified approach to quantifying algorithmic unfairness: Measuring
  individual \&group unfairness via inequality indices. In: Proceedings of the
  24th ACM SIGKDD International Conference on Knowledge Discovery \& Data
  Mining, pp 2239--2248

\bibitem[{Stapor et~al(2021)Stapor, Ksieniewicz, Garc{\'{\i}}a, and
  Wozniak}]{Stapor:2021}
Stapor K, Ksieniewicz P, Garc{\'{\i}}a S, Wozniak M (2021) How to design the
  fair experimental classifier evaluation. Appl Soft Comput 104:107219

\bibitem[{Subramanian et~al(2021)Subramanian, Rahimi, Baldwin, Cohn, and
  Frermann}]{subramanian2021fairness}
Subramanian S, Rahimi A, Baldwin T, Cohn T, Frermann L (2021) Fairness-aware
  class imbalanced learning. arXiv preprint arXiv:210910444

\bibitem[{Tao et~al(2019)Tao, Li, Guo, Ren, Li, Liu, and Zou}]{Tao:2019}
Tao X, Li Q, Guo W, Ren C, Li C, Liu R, Zou J (2019) Self-adaptive cost
  weights-based support vector machine cost-sensitive ensemble for imbalanced
  data classification. Inf Sci 487:31--56

\bibitem[{Wei et~al(2013)Wei, Li, Cao, Ou, and Chen}]{wei2013effective}
Wei W, Li J, Cao L, Ou Y, Chen J (2013) Effective detection of sophisticated
  online banking fraud on extremely imbalanced data. World Wide Web
  16(4):449--475

\bibitem[{Wen and Wu(2021)}]{Wen:2021}
Wen G, Wu K (2021) Building decision tree for imbalanced classification via
  deep reinforcement learning. In: Asian Conference on Machine Learning, {ACML}
  2021, 17-19 November 2021, Virtual Event, {PMLR}, Proceedings of Machine
  Learning Research, vol 157, pp 1645--1659

\bibitem[{Wo{\'z}niak et~al(2014)Wo{\'z}niak, Grana, and
  Corchado}]{wozniak2014survey}
Wo{\'z}niak M, Grana M, Corchado E (2014) A survey of multiple classifier
  systems as hybrid systems. Information Fusion 16:3--17

\bibitem[{Xu et~al(2021)Xu, Liu, Li, Jain, and Tang}]{Xu:2021}
Xu H, Liu X, Li Y, Jain AK, Tang J (2021) To be robust or to be fair: Towards
  fairness in adversarial training. In: Proceedings of the 38th International
  Conference on Machine Learning, {ICML} 2021, 18-24 July 2021, Virtual Event,
  {PMLR}, Proceedings of Machine Learning Research, vol 139, pp 11492--11501

\bibitem[{Yeom and Fredrikson(2020)}]{Yeom:2020}
Yeom S, Fredrikson M (2020) Individual fairness revisited: Transferring
  techniques from adversarial robustness. In: Proceedings of the Twenty-Ninth
  International Joint Conference on Artificial Intelligence, {IJCAI} 2020,
  ijcai.org, pp 437--443

\bibitem[{Yurochkin et~al(2019)Yurochkin, Bower, and
  Sun}]{yurochkin2019training}
Yurochkin M, Bower A, Sun Y (2019) Training individually fair ml models with
  sensitive subspace robustness. arXiv preprint arXiv:190700020

\bibitem[{Zafar et~al(2017)Zafar, Valera, Rogriguez, and
  Gummadi}]{zafar2017fairness}
Zafar MB, Valera I, Rogriguez MG, Gummadi KP (2017) Fairness constraints:
  Mechanisms for fair classification. In: Artificial Intelligence and
  Statistics, PMLR, pp 962--970

\bibitem[{Zemel et~al(2013)Zemel, Wu, Swersky, Pitassi, and
  Dwork}]{zemel2013learning}
Zemel R, Wu Y, Swersky K, Pitassi T, Dwork C (2013) Learning fair
  representations. In: International conference on machine learning, PMLR, pp
  325--333

\bibitem[{Zhang et~al(2018)Zhang, Lemoine, and Mitchell}]{zhang2018mitigating}
Zhang BH, Lemoine B, Mitchell M (2018) Mitigating unwanted biases with
  adversarial learning. In: Proceedings of the 2018 AAAI/ACM Conference on AI,
  Ethics, and Society, pp 335--340

\bibitem[{Zhang et~al(2019{\natexlab{a}})Zhang, Tan, Li, and
  Hong}]{ZhangT:2019}
Zhang C, Tan KC, Li H, Hong GS (2019{\natexlab{a}}) A cost-sensitive deep
  belief network for imbalanced classification. {IEEE} Trans Neural Networks
  Learn Syst 30(1):109--122

\bibitem[{Zhang et~al(2019{\natexlab{b}})Zhang, Ramezani, and
  Naeim}]{Zhang:2019wot}
Zhang W, Ramezani R, Naeim A (2019{\natexlab{b}}) Wotboost: Weighted
  oversampling technique in boosting for imbalanced learning. In: 2019 {IEEE}
  International Conference on Big Data {(IEEE} BigData), Los Angeles, CA, USA,
  December 9-12, 2019, {IEEE}, pp 2523--2531

\bibitem[{{\v{Z}}liobaite(2017)}]{vzliobaite2017measuring}
{\v{Z}}liobaite I (2017) Measuring discrimination in algorithmic decision
  making. Data Mining and Knowledge Discovery 31(4):1060--1089

\bibitem[{Zyblewski et~al(2021)Zyblewski, Sabourin, and
  Wozniak}]{Zyblewski:2021}
Zyblewski P, Sabourin R, Wozniak M (2021) Preprocessed dynamic classifier
  ensemble selection for highly imbalanced drifted data streams. Inf Fusion
  66:138--154

\end{thebibliography}

\end{document}